\documentclass[10pt]{article}
\usepackage{arxiv}
\usepackage{cite}
\usepackage[utf8]{inputenc}
\usepackage[T1]{fontenc}
\usepackage[protrusion=true,expansion=false]{microtype}
\usepackage{amsmath,amssymb,amsfonts}
\usepackage{graphicx}
\usepackage{booktabs}
\usepackage{multirow}
\usepackage{xcolor}
\usepackage{algorithm}
\usepackage{algpseudocode}
\usepackage{subcaption}
\usepackage{url}
\usepackage{bm}
\usepackage[colorlinks=true,linkcolor=blue,citecolor=blue,urlcolor=blue]{hyperref}

\newcommand{\orcidicon}[1]{\href{https://orcid.org/#1}{\includegraphics[width=0.32cm]{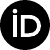}}}

\def\BibTeX{{\rm B\kern-.05em{\sc i\kern-.025em b}\kern-.08em
    T\kern-.1667em\lower.7ex\hbox{E}\kern-.125emX}}

\newcommand{\stmoe}{\textsc{ST-MoE}}
\newcommand{\dmodel}{d_\text{model}}
\newcommand{\dspace}{d_\text{space}}
\newcommand{\topk}{\text{top-}k}

\newcommand{\eg}{\textit{e.g.}}

\newcommand{\Paragraph}[1]{\vspace{0.5em}\noindent\textbf{#1}}

\title{Equifinality in Mixture of Experts: \\
Routing Topology Does Not Determine Language Modeling Quality}

\author{
  Ivan Ternovtsii \orcidicon{0009-0009-9267-8516}\thanks{This research was conducted as part of PhD studies at the Department of Software Systems, Faculty of Information Technologies, Uzhhorod National University. HengeBytes generously provided computational resources. We thank the Department of Software Systems for academic support and the HengeBytes team for maintaining the computational infrastructure. This paper reports 62 controlled experiments totaling approximately 460 GPU-hours. Corresponding author: Ivan Ternovtsii (e-mail: ivan.ternovtsii@uzhnu.edu.ua).} \\
  Department of Software Systems, Uzhhorod National University\\
  Narodna sq. 3, Uzhhorod, Ukraine, 88000\\
  HengeBytes\\
  \texttt{ivan.ternovtsii@uzhnu.edu.ua} \\
  \And
  Yurii Bilak \orcidicon{0000-0001-5989-1643} \\
  Department of Software Systems, Uzhhorod National University\\
  Narodna sq. 3, Uzhhorod, Ukraine, 88000\\
}

\date{April 2026}

\begin{document}
\maketitle

\begin{abstract}
Sparse Mixture-of-Experts (MoE) architectures employ increasingly sophisticated routing mechanisms---learned routers, multi-hop trajectories, token-dependent gating.
We ask: \emph{does routing topology actually determine language modeling quality?}

We build a \textbf{geometric MoE} (\stmoe{}) using cosine-similarity routing against learned centroids in a low-dimensional space ($\dspace = 64$), requiring \textbf{80\% fewer routing parameters} than standard linear routers.
Through 62 controlled experiments on WikiText-103 at 76--84M parameters trained to convergence (50K steps, 1.64B tokens), we find that \textbf{routing topology does not determine asymptotic perplexity (PPL)}: five cosine-routing variants are statistically equivalent within a 1-PPL margin (Two One-Sided Tests [TOST], $p < 0.05$ for all 10 pairwise comparisons; 15 runs across 3 seeds, observed range 33.93--34.72).
The finding extends to hash, random-fixed, and top-1 routing (single-seed; graceful 1.1--2.2~PPL degradation) and replicates on OpenWebText (0.03~PPL gap, 6 runs, 3 seeds each).
A standard linear router with $5.3\times$ more routing parameters reaches PPL~32.76, but iso-parameter cosine routing closes 67\% of this gap---the true mechanism advantage is ${\sim}1.2\%$.

The mechanistic explanation is convergent redundancy: multi-hop updates are collinear ($\cos(\Delta h_0, \Delta h_1) = 0.805$), implementing magnitude amplification rather than compositional reasoning; a single learnable scalar replicates multi-hop performance.
As a practical payoff, zero-shot relative-norm halting saves 25\% of MoE FLOPs at $+0.12\%$ PPL.
Expert-level specialization and causal controllability---which coexist with topology-level equifinality---are explored in a companion paper \cite{controllability2026}.
\end{abstract}


\section{Introduction}
\label{sec:intro}

Sparse Mixture-of-Experts (MoE) models achieve remarkable efficiency by activating only a subset of parameters per token \cite{shazeer2017outrageously, lepikhin2021gshard, fedus2022switch}.
The standard paradigm employs a \emph{learned router}---a small neural network that maps token representations to expert assignment probabilities.
This design introduces several well-known challenges: load imbalancing requires auxiliary losses \cite{fedus2022switch}, expert collapse demands careful initialization \cite{zoph2022stmoe}, and the router itself constitutes an opaque bottleneck whose learned representations may diverge from the semantic content it routes \cite{liu2023janus}.

Meanwhile, the MoE community has invested heavily in routing topology: multi-hop trajectories, hierarchical gating, chain-of-experts composition.
The implicit assumption is that more sophisticated routing produces better models.
\textbf{We test this assumption directly.}

Our tool is a \emph{geometric MoE} (\stmoe{}) in which each expert is associated with a learned centroid vector.
Routing is computed as cosine similarity between the token's projected representation and expert centroids in a low-dimensional space ($\dspace = 64$).
This requires 1.57M routing parameters (projections + centroids) vs.\ 8.39M for a standard \texttt{nn.Linear} router---an \textbf{80\% reduction}.
The architecture also supports \emph{multi-hop re-routing}: after each expert update, the token re-projects its updated state and routes again, creating a trajectory through expert space.

Through 62 controlled experiments culminating in a convergence study at 76M parameters on 1.64B tokens, we establish the following:

\begin{enumerate}
    \item \textbf{Routing topology equifinality} (Section~\ref{sec:convergence}): Five cosine-routing variants converge to statistically equivalent PPL (TOST, $p < 0.05$ for all 10 pairwise comparisons; 15 runs across 3 seeds, range 33.93--34.72).
    The remaining gap to a standard linear router (PPL~32.76, $5.3\times$ more routing parameters) is mostly explained by routing \emph{capacity}, not mechanism or topology.
    The hierarchy: routing capacity $>$ routing mechanism $>$ routing topology $\approx$ composition method $\gg$ expert capacity.

    \item \textbf{Mechanistic explanation} (Section~\ref{sec:mechanism}): Multi-hop updates are collinear ($\cos = 0.805$), performing convergent magnitude amplification rather than compositional reasoning.
    A single learnable scalar replicates multi-hop performance.
    Cross-seed analysis reveals $500\times$-above-random functional overlap through entirely different weight parameterizations---the quantitative signature of equifinality.

    \item \textbf{Practical payoff} (Section~\ref{sec:halting}): Since routing topology is quality-neutral, we exploit geometric routing for compute savings.
    Zero-shot relative-norm halting saves 25\% of MoE FLOPs at $+0.12\%$ PPL, validated at convergence.
    Architecture dualism (Section~\ref{sec:ablations}) further reveals that multi-hop requires many tiny experts while single-hop favors fewer large ones---governed by centroid density in routing space.
\end{enumerate}

\section{Background}
\label{sec:background}

\Paragraph{Standard MoE routing.}
In a typical sparse MoE layer \cite{shazeer2017outrageously, fedus2022switch}, the router is a learned linear projection $W_r \in \mathbb{R}^{\dmodel \times M}$ mapping each token representation to $M$ expert logits.
After softmax, top-$k$ experts are selected and their outputs combined with the routing weights.
For $M = 1024$ experts and $\dmodel = 1024$, this router requires $M \times \dmodel = 1{,}048{,}576$ parameters per layer.

\Paragraph{Load balancing.}
Without regularization, learned routers collapse to a few experts.
The standard remedy is an auxiliary balance loss $\mathcal{L}_\text{bal} = \alpha \cdot M \sum_{i=1}^M f_i p_i$ \cite{fedus2022switch}, where $f_i$ is the fraction of tokens routed to expert $i$ and $p_i$ is the mean routing probability.

\Paragraph{Routing parameter budget.}
The router is often treated as negligible overhead.
However, at fine granularity ($M \geq 1024$), routing parameters become substantial: 8 layers $\times \dmodel \times M = 8.39$M for our configuration---10\% of total model parameters.
Our cosine routing reduces this to 1.57M (80\% less) via a low-dimensional bottleneck, with consequences we quantify in Section~\ref{sec:convergence}.

\section{Method: Semantic Trajectory MoE}
\label{sec:method}

\subsection{Architecture Overview}

\begin{figure}[t]
    \centering
    \includegraphics[width=\textwidth]{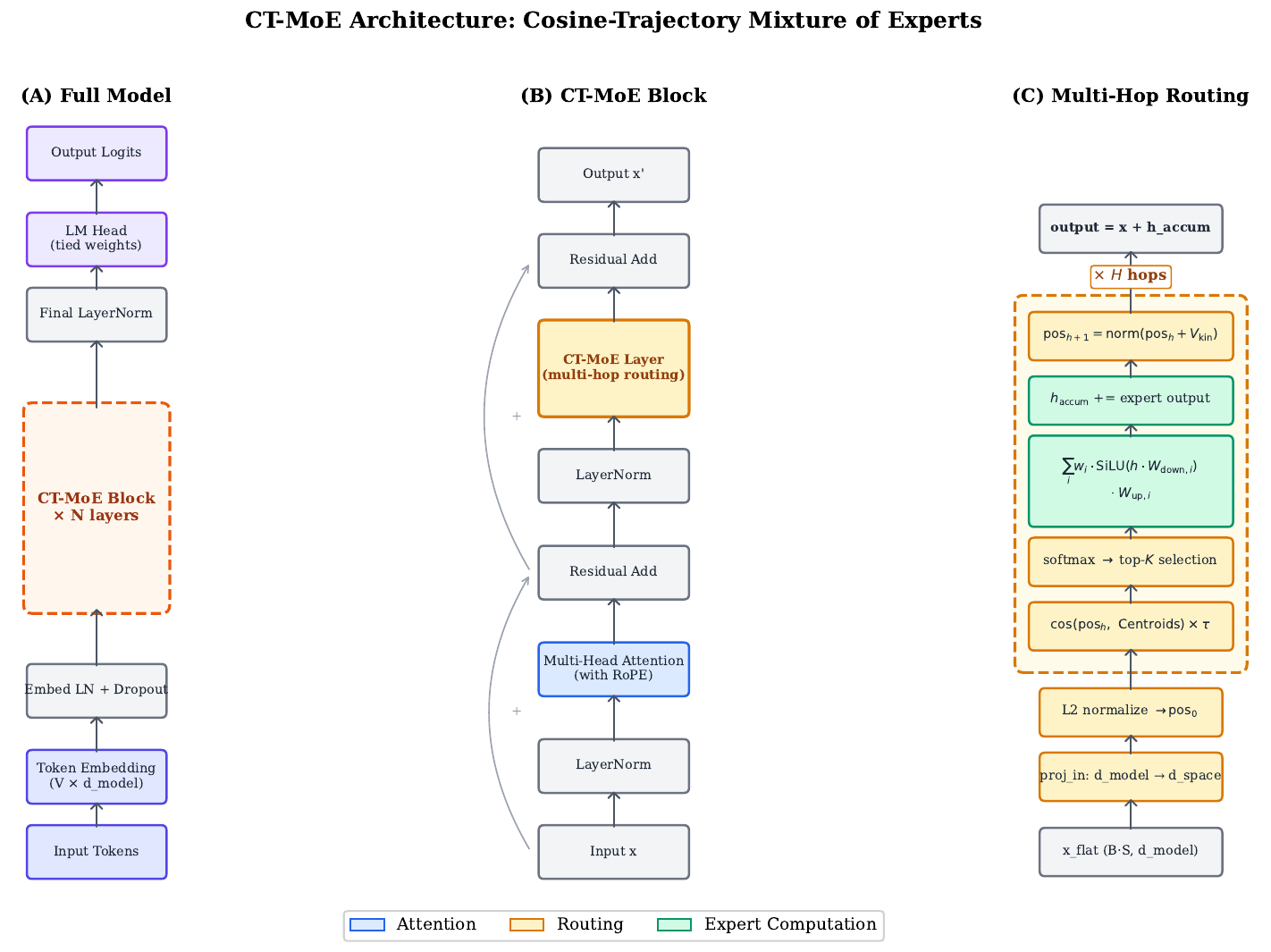}
    \caption{\textbf{\stmoe{} Architecture.} (A)~Full model: token embedding $\to$ $N$ blocks $\to$ LM head. (B)~Single block: pre-LN attention with RoPE, followed by pre-LN \stmoe{} layer, both with residual connections. (C)~Multi-hop routing: tokens are projected into a $\dspace$-dimensional coordinate space, routed to top-$K$ experts via cosine similarity with learned centroids, and accumulate expert updates across $H$ hops with semantic position updates between hops.}
    \label{fig:architecture}
\end{figure}

\stmoe{} is a pre-LayerNorm Transformer where each block consists of:
\begin{equation}
    x \leftarrow x + \text{MHA}(\text{LN}(x)), \qquad
    x \leftarrow x + \text{ST-MoE}(\text{LN}(x))
\end{equation}
where MHA uses multi-head attention with Rotary Position Embeddings \cite{su2024roformer}, and ST-MoE replaces the dense feed-forward network (FFN) with our sparse multi-hop routing layer.
Weight tying between embedding and LM head \cite{press2017using} is used throughout.
Figure~\ref{fig:architecture} provides a schematic overview.

\subsection{Geometric Cosine Routing}

Given a token representation $h \in \mathbb{R}^{\dmodel}$, we project it into a lower-dimensional coordinate space:
\begin{equation}
    \text{pos} = \text{normalize}\!\left(\text{proj}_\text{in}(h)\right) \in \mathbb{R}^{\dspace}
\end{equation}
where $\text{proj}_\text{in}: \mathbb{R}^{\dmodel} \to \mathbb{R}^{\dspace}$ is a linear projection (no bias) and normalization is L2.
Each of the $M$ experts has a learned centroid $c_i \in \mathbb{R}^{\dspace}$, also L2-normalized.
Routing scores are:
\begin{equation}
    s_i = \tau \cdot \langle \text{pos},\, c_i \rangle, \qquad
    p_i = \frac{\exp(s_i)}{\sum_{j=1}^M \exp(s_j)}
\end{equation}
where $\tau = 30$ is a fixed cosine scale \cite{zhai2023sigmoid}.
The top-$k$ experts are selected from $p$, with renormalized weights $w_i = p_i / \sum_{j \in \topk} p_j$.

\Paragraph{Routing parameter budget.}
This routing mechanism requires $\text{proj}_\text{in}$ $(\dmodel \times \dspace)$, centroids $C$ $(M \times \dspace)$, and optional kinematic vectors---totaling 1.57M parameters across 8 layers in our largest configuration.
A standard linear router (\texttt{nn.Linear($\dmodel$, $M$)}) would require 8.39M---a $5.3\times$ difference.
We do \emph{not} claim ``zero routing parameters''; we claim an 80\% reduction enabled by the low-dimensional bottleneck.

\Paragraph{Why $\dspace = 64$?}
The projection bottleneck concentrates routing signal.
In our ablation (Exp~004), routing directly in $\dmodel = 768$ space degraded PPL by 3.1\%, with multi-hop trajectories degenerating to repeated expert selection.

\Paragraph{Semantic re-routing.}
Once tokens have been processed by multi-head attention, their semantic state is a sufficient compass.
Explicit navigation vectors (kinematic updates to routing position) provide no benefit when attention is present (Exp~002: $\Delta < 2.5$ PPL across all routing modes).
We use \emph{semantic re-routing}: after each hop, position is recomputed from the token's updated semantic state $\text{pos} \leftarrow \text{normalize}(\text{proj}_\text{in}(x + h_\text{accum}))$.
This requires \emph{no additional navigation parameters} beyond the shared $\text{proj}_\text{in}$---the token's updated meaning determines its routing position.

\subsection{Expert Parameterizations}

\Paragraph{Static experts} ($V_\text{sem}$): Each expert stores a learned vector $V_i \in \mathbb{R}^{\dmodel}$.
The update is a weighted sum $\Delta h = \sum_{i \in \topk} w_i V_i$.
This is the original formulation.
Its critical limitation is \emph{linear composition}: multi-hop updates collapse to $h + \sum V_k$ regardless of trajectory.

\Paragraph{MLP experts} (rank-$r$ nonlinear): Each expert stores $W_{\text{down},i} \in \mathbb{R}^{r \times \dmodel}$, $W_{\text{up},i} \in \mathbb{R}^{\dmodel \times r}$:
\begin{equation}
    \Delta h = \sum_{i \in \topk} w_i \cdot W_{\text{up},i}\, \text{SiLU}\!\left(W_{\text{down},i}\, h_\text{current}\right)
    \label{eq:mlp_expert}
\end{equation}
where $h_\text{current} = x + h_\text{accum}$ is the token's accumulated state and $r$ is the expert size (rank).
The nonlinearity breaks the linear trap: each hop's update depends on the \emph{current} state, enabling genuine recursive composition $f_3(f_2(f_1(x)))$.
For singleton experts ($r = 1$), this reduces to $\text{SiLU}(h \cdot w_\text{down}) \cdot w_\text{up}$.

\subsection{Multi-Hop Semantic Trajectory}
\label{sec:method_trajectory}

The defining feature of \stmoe{} is iterative re-routing:

\begin{algorithm}[H]
\caption{ST-MoE Forward Pass. Each token is projected into routing space, matched to top-$K$ experts via cosine similarity, and updated over $H$ hops with semantic re-routing between hops.}
\begin{algorithmic}[1]
\State $h_\text{accum} \leftarrow \mathbf{0}$
\State $\text{pos} \leftarrow \text{normalize}(\text{proj}_\text{in}(x))$ \Comment{Initial position}
\For{$\text{hop} = 1, \ldots, H$}
    \State $\{i_1, \ldots, i_k\}, \{w_1, \ldots, w_k\} \leftarrow \topk\text{-route}(\text{pos}, C)$ \Comment{Nearest experts}
    \State $\Delta h \leftarrow \text{expert\_update}(\{i\}, \{w\}, x + h_\text{accum})$ \Comment{Nonlinear update}
    \State $h_\text{accum} \leftarrow h_\text{accum} + \Delta h$ \Comment{Accumulate}
    \If{$\text{hop} < H$}
        \State $\text{pos} \leftarrow \text{normalize}(\text{proj}_\text{in}(x + h_\text{accum}))$ \Comment{Semantic re-route}
    \EndIf
\EndFor
\State \Return $x + h_\text{accum}$
\end{algorithmic}
\end{algorithm}

After each hop, the token's routing position is \emph{recomputed} from its updated semantic state.
Experts are shared across all hops---the same expert can serve different roles depending on where a token's trajectory brings it.
In our ablation (Exp~016), shared experts (PPL~302.54) beat iso-parameter unrolled per-hop pools (PPL~310.72) by $-2.6\%$, with unrolled pools exhibiting routing collapse to identical expert selection at every hop.

\subsection{Training}

We use the standard Switch Transformer balance loss \cite{fedus2022switch}:
\begin{equation}
    \mathcal{L} = \mathcal{L}_\text{LM} + \alpha \cdot M \sum_{i=1}^M f_i p_i
\end{equation}
with $\alpha = 0.05$, where $f_i$ is the fraction of tokens routed to expert $i$ and $p_i$ is the mean routing probability.
Cosine learning rate schedule with 200-step warmup, AdamW with $(\beta_1, \beta_2) = (0.9, 0.95)$.

\subsection{Experimental Setup}
\label{sec:setup}

All experiments use WikiText-103 \cite{merity2017pointer} with a 32K Byte Pair Encoding (BPE) vocabulary across three configurations at two model sizes:

\begin{table}[h]
\centering
\caption{Model configurations used throughout the paper, varying scale, depth, expert count, and training duration.}
\label{tab:configs}
\begin{tabular}{lccccccc}
\toprule
Scale & $\dmodel$ & Heads & Layers & $M$ & Steps & Tokens & Params \\
\midrule
Small (36M) & 768 & 12 & 4 & 512 & 5,000 & 82M & 37.6M \\
Large (76M) & 1024 & 16 & 6 & 1024 & 20,000 & 328M & 76--84M \\
Marathon (76M) & 1024 & 16 & 8 & 1024 & 50,000 & 1.64B & 76--84M \\
\bottomrule
\end{tabular}
\end{table}

Default: $\dspace = 64$, $\topk = 4$, $H = 3$ hops, expert\_size $= 1$ (singleton), semantic re-route, $\tau = 30$, bfloat16.
All iso-FLOP comparisons match active neurons per token.
Seed $= 42$, NVIDIA RTX 3090 GPUs.
The Marathon configuration trains to convergence: 50K steps $\times$ 32,768 tokens/step $= 1.64$B tokens ($\sim$14 epochs of WikiText-103), following Chinchilla scaling \cite{hoffmann2022training} to ensure 76M parameters are fully trained.

\subsection{Reproducibility}
\label{sec:reproducibility}

All experiments use the same codebase, data pipeline, and hardware.
\textbf{Compute budget}: 62 experiments totaling approximately 460 GPU-hours on NVIDIA RTX 3090 (24~GB) and RTX 3090~Ti (24~GB) GPUs; the convergence marathon (Exps~025--027) accounts for $\sim$20h, multi-seed validation (Exps~042--043, 049) $\sim$72h, routing interventions (Exps~033--035) $\sim$24h, and remaining experiments $\sim$344h cumulatively.
\textbf{Seed strategy}: primary seed $= 42$ for all experiments; multi-seed validation uses seeds 42, 137, and 7 for all five cosine-routing variants ($5 \times 3 = 15$ runs); statistical validation via paired bootstrap with 10{,}000 resamples.
\textbf{Data}: WikiText-103 \cite{merity2017pointer} tokenized with a 32K BPE vocabulary; all models see the same token sequences in the same order (given the same seed).
\textbf{Software}: PyTorch 2.0+, bfloat16 mixed precision, AdamW optimizer with $(\beta_1, \beta_2) = (0.9, 0.95)$.
Code and checkpoints will be released upon publication.

\section{Main Results}
\label{sec:results}

\Paragraph{Underfitting-regime results.}
During early training (5--20K steps), depth provides substantial advantages: rank-1 MLP experts improve PPL by $-4.4\%$ over static experts, and the depth advantage grows with scale (up to $+8.8\%$ at 76M parameters).
However, these results are \emph{superseded} by the convergence results below, where the depth advantage inverts---Wide beats Deep by $-2.0\%$.
We present underfitting results as mechanistic characterization in Appendix~\ref{app:underfitting}.

\subsection{Iso-FLOP Comparison with Dense Baselines}

We first compare MoE against dense baselines at iso-FLOP---matching active computation per token.
These baselines have very narrow FFNs ($d_\text{ff} = 12$ or $48$), which no practitioner would deploy; the purpose is to isolate the effect of \emph{conditional computation}.
Section~\ref{sec:dense_iso_param} presents the proper iso-parameter comparison.

\begin{table}[h]
\centering
\caption{MoE vs Dense at iso-FLOP (same active computation per token).}
\label{tab:dense_comparison}
\begin{tabular}{lccc}
\toprule
Model & Active/Token & PPL & vs Dense \\
\midrule
\multicolumn{4}{c}{\emph{Small active budget (12 neurons)}} \\
Dense d\_ff=12 & 12 & 306.01 & baseline \\
Singleton MoE ($3 \times 4 \times 1$) & 12 & 304.50 & $-0.5\%$ \\
\midrule
\multicolumn{4}{c}{\emph{Medium active budget (48 neurons)}} \\
Dense d\_ff=48 & 48 & 302.24 & baseline \\
Block MoE Wide ($1 \times 6 \times 8$) & 48 & \textbf{298.91} & $\mathbf{-1.1\%}$ \\
\bottomrule
\end{tabular}
\end{table}

At both budget levels, sparse MoE matches or beats dense.
The block MoE advantage ($-1.1\%$) demonstrates a genuine quality gain from conditional computation: each token assembles a customized transformation from $\binom{64}{6} \approx 75$M possible expert combinations.
Note that the winning MoE configuration at 48~active neurons is \emph{Wide} (single-hop, block experts)---a point we return to in Section~\ref{sec:dualism}.

\subsection{Iso-Parameter Dense Baseline}
\label{sec:dense_iso_param}

The iso-FLOP comparison above is the practically relevant one: MoE's purpose is to scale total parameters while fixing active computation per token, exactly as in production systems like Mixtral and DeepSeek-V3.
However, for completeness, we also train a dense transformer with $d_\text{ff} = 1120$, matching the exact parameter count of our Wide 1$\times$12 MoE (84,742,144 parameters, verified by assertion).
This model uses ${\sim}93\times$ more active FFN computation per token than the MoE ($d_\text{ff}{=}1120$ vs.\ 12 active rank-1 experts).

The dense model converges to PPL~\textbf{29.36}, significantly outperforming all MoE variants.
This is expected: with $93\times$ more active computation, a dense FFN should win.
The result confirms the standard MoE motivation---sparse expert selection achieves competitive quality (33.93 vs.\ 34.49 at iso-FLOP) while using only a fraction of the total parameters per token, enabling parameter scaling beyond what dense models can afford to activate at inference time.

\begin{figure}[t]
    \centering
    \includegraphics[width=\textwidth]{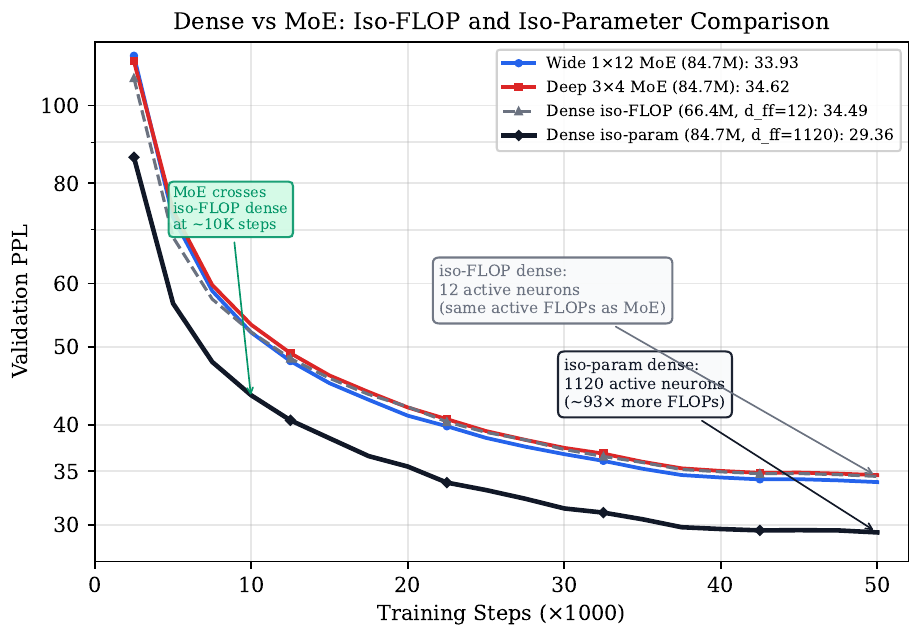}
    \caption{\textbf{Dense vs MoE Training Comparison.} The iso-FLOP dense baseline ($d_\text{ff}{=}12$, same active FLOPs) tracks MoE throughout training; the iso-parameter dense baseline ($d_\text{ff}{=}1120$, ${\sim}93\times$ more active FLOPs) dominates after step 5K. MoE crosses the iso-FLOP dense at ${\sim}$10K steps, confirming that sparse routing provides value at matched active computation.}
    \label{fig:dense_vs_moe}
\end{figure}

It also reveals a secondary finding: rank-1 experts provide individually decodable units, but their expressivity per active parameter is lower than a full-rank FFN.
At this small scale, the interpretability benefit of rank-1 structure comes at a per-parameter efficiency cost.
Fine-grained MoE scaling laws \cite{ludziejewski2024scaling} predict that this gap narrows with scale: finer granularity ($G = d_\text{ff}/d_\text{expert}$) yields diminishing but persistent gains following $G^{-0.58}$, and our rank-1 experts represent the extreme of this continuum.
Whether billion-parameter scale closes the remaining gap remains an open question.
Figure~\ref{fig:dense_vs_moe} visualizes the training dynamics of all four baselines.

\subsection{The Central Result: Routing Topology Equifinality}
\label{sec:convergence}

The preceding results (PPL~136--320 at 5--20K steps) are obtained during underfitting.
To establish asymptotic behavior, we train all configurations to convergence: 50K steps on 1.64B tokens ($\sim$14 epochs), following Chinchilla scaling \cite{hoffmann2022training}.
The result is our central finding.

\begin{figure}[t]
    \centering
    \includegraphics[width=\textwidth]{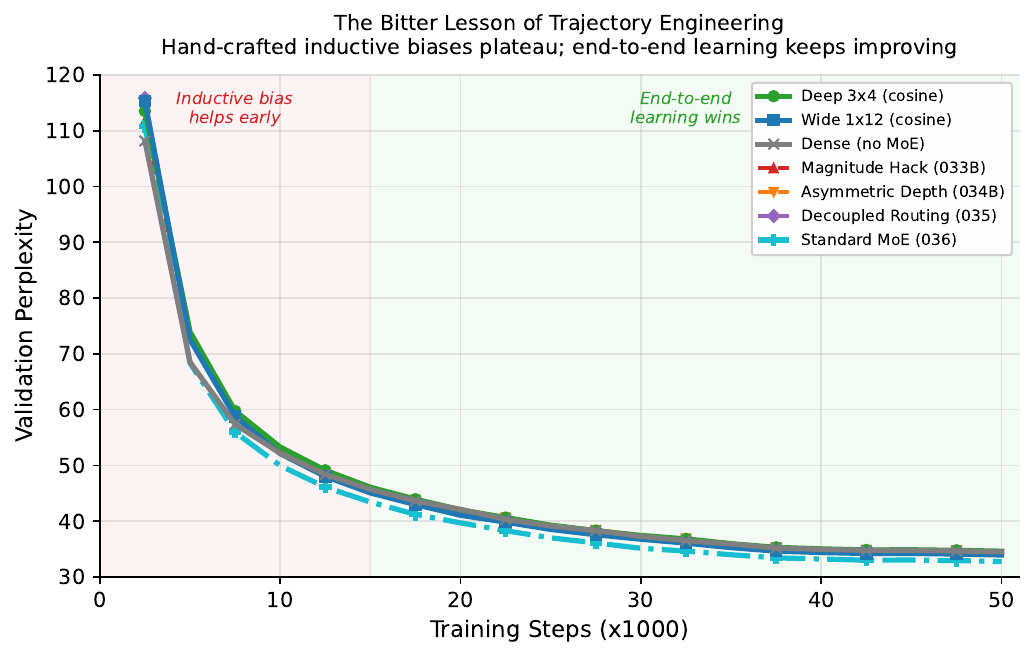}
    \caption{\textbf{Convergence Marathon.} Training curves for all configurations at 76M parameters, 50K steps (1.64B tokens). Among cosine-routing variants, all topologies converge to a statistically equivalent band (TOST, $\delta{=}1$~PPL, all pairs $p < 0.05$; observed range 33.93--34.72 across 15 runs). The standard linear router (dashed cyan) falls outside this band at PPL~32.76, demonstrating that routing \emph{capacity} matters even when routing \emph{topology} does not.}
    \label{fig:convergence}
\end{figure}

\begin{figure}[t]
    \centering
    \includegraphics[width=\textwidth]{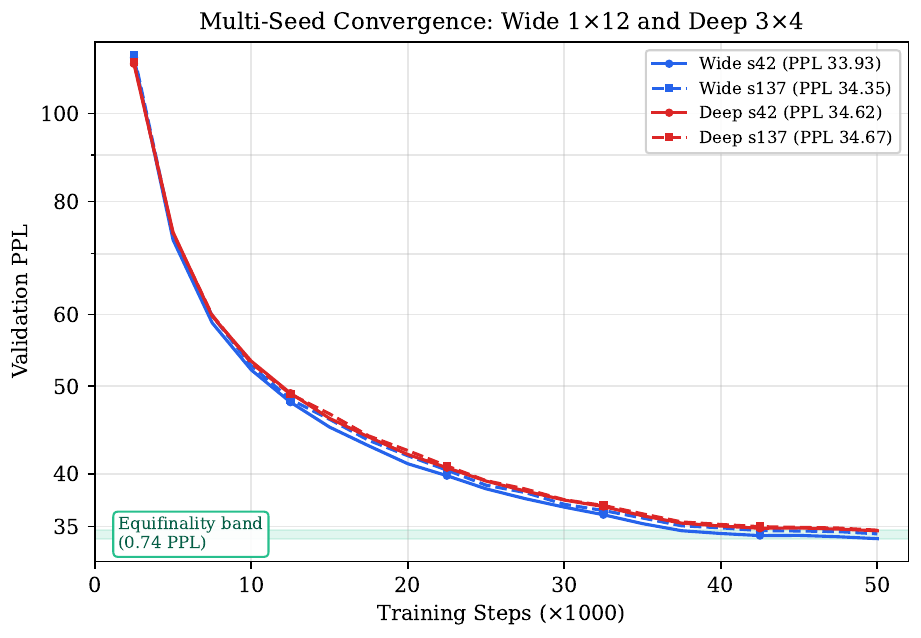}
    \caption{\textbf{Multi-Seed Validation.} Training curves for Wide 1$\times$12 and Deep 3$\times$4 with seeds 42 and 137. Same-topology curves track nearly identically throughout training: Wide $\Delta{=}1.2\%$, Deep $\Delta{=}0.1\%$. All four runs converge within a 0.74~PPL band. Full 5-variant $\times$ 3-seed validation (15 runs total) confirms all topologies converge within a 0.79~PPL band (33.93--34.72), with inter-variant spread ($0.60$~PPL) only $5.0\times$ the average seed noise ($0.12$~PPL avg std).}
    \label{fig:cross_seed}
\end{figure}

\begin{table}[t]
\centering
\caption{Convergence results at 76--84.7M parameters, 1.64B tokens (RTX 3090). Among cosine-routing variants, all topologies are statistically equivalent within 1~PPL (TOST, $p < 0.05$ for all 10 pairwise comparisons). Mean$\pm$Std computed over 3 seeds (42, 137, 7) for each cosine variant; all 15 runs fall within a 0.79~PPL band. Broader topology conditions (Exp~065) and rank-16 experts (Exp~070) extend the comparison. MoE matches the iso-FLOP dense baseline ($d_\text{ff}{=}12$) while using the same active computation but ${\sim}93\times$ more total parameters. The iso-parameter dense baseline ($d_\text{ff}{=}1120$, $93\times$ more active FLOPs) outperforms all MoE variants, confirming MoE's value is conditional computation, not per-parameter efficiency.}
\label{tab:convergence}
\resizebox{\textwidth}{!}{%
\begin{tabular}{lcccccc}
\toprule
Model & Architecture & Final PPL & Mean$\pm$Std (3 seeds) & vs iso-FLOP Dense & Routing Params & Train Time \\
\midrule
\multicolumn{7}{c}{\emph{Baselines}} \\
Dense iso-FLOP & d\_ff=12 & 34.49 & --- & --- & --- & 3.5h \\
Dense iso-param & d\_ff=1120 ($93\times$ active) & \textbf{29.36} & --- & $-14.9\%$ & --- & 3.9h \\
\stmoe{} Wide 1$\times$12 & cosine, 1 hop, top-12 & 33.93 & $34.22 \pm 0.25$ & $-1.6\%$ & 1.57M & 7.4h \\
\stmoe{} Deep 3$\times$4 & cosine, 3 hops, top-4 & 34.62 & $34.65 \pm 0.03$ & $+0.4\%$ & 1.57M & 8.9h \\
\midrule
\multicolumn{7}{c}{\emph{Cosine Routing Interventions (Exps~033--035)}} \\
Decoupled Routing (035) & per-hop proj\_in, 3$\times$4 & 34.05 & $34.05 \pm 0.10$ & $-1.3\%$ & 2.10M & 9.5h \\
Magnitude Hack (033B) & Wide + learned $\alpha$ & 34.11 & $34.10 \pm 0.08$ & $-1.1\%$ & 1.57M & 7.0h \\
Asymmetric Depth (034B) & [2,1,2,1,1,4,4,2] hops & 34.59 & $34.60 \pm 0.12$ & $+0.3\%$ & 1.57M & 7.7h \\
\midrule
\multicolumn{7}{c}{\emph{Iso-Parameter Controls (Exps~036--036C)}} \\
Standard MoE Wide (036) & \texttt{nn.Linear}, 1 hop, top-12 & \textbf{32.76} & --- & $\mathbf{-5.0\%}$ & 8.39M & 7.5h \\
Cosine d=341 (036B) & cosine, $\dspace = 341$, 1 hop, top-12 & 33.14 & --- & $-3.9\%$ & 8.38M & 7.6h \\
Bigger Experts (036C) & cosine, $\dspace = 64$, expert\_size=2 & 34.92 & --- & $+1.2\%$ & 1.57M & 11.6h \\
\midrule
\multicolumn{7}{c}{\emph{Broader Topology (Exp~065)}} \\
Hash Routing & deterministic, no learning & 36.04 & --- & $+4.6\%$ & 0 & 4.5h \\
Random-Fixed Routing & frozen random centroids & 35.03 & --- & $+1.6\%$ & 1.05M (frozen) & 5.4h \\
Top-1 Cosine & winner-take-all, $K{=}1$ & 36.14 & --- & $+4.8\%$ & 1.05M & 5.1h \\
\midrule
\multicolumn{7}{c}{\emph{Rank-16 Experts (Exp~070; 256 experts, 4 layers, $K{=}4$)}} \\
Wide 1$\times$4 rank-16 & cosine, 1 hop, top-4 & 37.77 & $37.72 \pm 0.08$ & $+9.4\%$ & 0.33M & 10.9h \\
Deep 2$\times$2 rank-16 & cosine, 2 hops, top-2 & 38.30 & $38.27 \pm 0.05$ & $+11.0\%$ & 0.33M & 11.9h \\
\bottomrule
\end{tabular}%
}
\end{table}

\Paragraph{Topology equifinality.}
The five cosine-routing configurations converge within a \textbf{0.79~PPL band} (33.93--34.72 across 15 runs: 5 variants $\times$ 3 seeds).
To validate this statistically, we apply the Two One-Sided Tests (TOST) equivalence framework \cite{schuirmann1987comparison} using paired bootstrap confidence intervals (10{,}000 resamples over 50 validation batches).
At an equivalence margin of $\delta = 0.03$ cross-entropy loss ($\approx$1~PPL), all 10 pairwise comparisons among the five cosine variants pass TOST ($p < 0.05$), confirming formal equivalence:

\begin{equation}
    \forall\, (i,j) \in \binom{5}{2}: \quad |\bar{\ell}_i - \bar{\ell}_j| < \delta = 0.03 \quad \text{(TOST, } p < 0.05\text{)}
    \label{eq:tost}
\end{equation}

The maximum observed pairwise difference is 0.023 cross-entropy loss ($\approx$0.75~PPL, between Wide and Deep), which is $1.8\times$ the intra-seed variance (0.013 loss for Wide across seeds 42 vs.\ 137).
Paired bootstrap CIs on loss differences are tight (typical width $\pm$0.008 loss), with 4 of 10 pairs including zero and 6 showing small but statistically detectable differences---all within the 1-PPL equivalence margin.
Full multi-seed validation (3 seeds $\times$ 5 variants $= 15$ runs) confirms equifinality across all topologies:
Wide reproduces at 33.93/34.35/34.37 (std$=$0.25), Deep at 34.62/34.67/34.67 (std$=$0.03), Magnitude Hack at 34.11/34.17/34.02 (std$=$0.08), Asymmetric at 34.59/34.49/34.72 (std$=$0.12), and Decoupled at 34.05/33.95/34.14 (std$=$0.10).
The inter-variant spread of means (0.60~PPL) is $5.0\times$ the average intra-variant standard deviation (0.12~PPL) and $2.4\times$ the largest variant's seed noise (Wide, $\sigma = 0.25$~PPL), confirming that topology effects are modest relative to seed variance.

The iso-FLOP dense baseline ($d_\text{ff}{=}12$) sits at 34.49---in the middle of this band.
The linear router, serving as a \emph{negative} equivalence control, falls clearly outside: its mean loss is 0.033 below the best cosine variant ($p < 0.001$, paired bootstrap), confirming the equivalence band is specific to cosine routing topologies, not an artifact of insensitive measurement.
An iso-parameter dense baseline ($d_\text{ff}{=}1120$, same 84.7M parameters but ${\sim}93\times$ more active computation) reaches PPL~29.36 (Section~\ref{sec:dense_iso_param}), confirming that MoE's value is conditional computation---scaling parameters while fixing active FLOPs---not per-parameter efficiency.

\Paragraph{Broader routing spectrum.}
Exp~065 extends the topology comparison beyond learned cosine routing to include hash routing (deterministic position-based assignment, no learning), random-fixed routing (frozen random centroids), and top-1 winner-take-all routing (all single-seed; multi-seed confirmation is left to future work).
Hash and random-fixed routing achieve PPL~36.04 and 35.03 respectively---only 1.1--2.1~PPL worse than the best learned routing (33.93), and both gaps are $4$--$9\sigma$ above seed noise ($\sigma = 0.25$~PPL).
Even without learned routing, expert parameters alone learn useful representations.
Top-1 routing (PPL~36.14) shows that reducing $K$ from 12 to 1 costs roughly the same as removing learned routing entirely.
The full spectrum from no-routing to learned routing shows graceful degradation rather than a sharp transition: the gap between ``no routing'' and ``learned routing'' (1.1--2.2~PPL) is only ${\sim}2$--$3\times$ the spread within learned routing (0.79~PPL).

\Paragraph{Routing capacity breaks the band.}
The standard linear router (Exp~036) achieves PPL~\textbf{32.76}---1.17~PPL below the best cosine variant (33.93).
However, the linear router uses $5.3\times$ more routing parameters (8.39M vs.\ 1.57M), confounding mechanism with capacity.

Iso-parameter controls disentangle this.
Expanding cosine routing's bottleneck from $\dspace = 64$ to $\dspace = 341$ to match the linear router's 8.38M routing parameters yields PPL~\textbf{33.14} (Exp~036B)---closing \textbf{67\%} of the gap (1.17 $\to$ 0.38~PPL).
Conversely, doubling expert MLP width (\texttt{expert\_size=2}, Exp~036C) adds 17M parameters to experts but \emph{degrades} PPL to 34.92---worse than the baseline despite 20\% more total parameters.

This reveals a refined hierarchy: \emph{routing capacity} $>$ \emph{routing mechanism} (linear vs.\ cosine at iso-params: 0.38~PPL, 1.2\%) $>$ \emph{routing topology} (0.79~PPL band across 15 runs, formally equivalent by TOST, Eq.~\ref{eq:tost}) $\approx$ \emph{composition method} (all variants worse; Section~\ref{sec:composition}) $\gg$ \emph{expert capacity} (negative returns).
The original 3.5\% gap was mostly a parameter count artifact; the true mechanism advantage of linear over cosine routing is $\sim$1.2\%.

\Paragraph{Training cost.}
Training times on a single RTX 3090 (Table~\ref{tab:convergence}) further underscore the expert capacity finding: bigger experts (036C) require \textbf{57\% more wall-clock time} (11.6h vs.\ 7.4h) due to doubled expert MLP width, while \emph{degrading} PPL by 2.9\%.
Dense training is fastest (3.5--3.9h) because it bypasses routing entirely; among MoE variants, Wide (7.4h) is faster than Deep (8.9h) due to fewer sequential hops.
Decoupled routing (035, 9.5h) pays a modest cost for per-hop projections.
Notably, the standard linear router (036, 7.5h) matches cosine Wide (7.4h) in training time despite $5.3\times$ more routing parameters---routing overhead is small relative to expert computation.

\Paragraph{Routing interventions confirm equifinality.}
To stress-test this finding, we train three deliberate routing interventions to 50K steps with matched LR schedules:
\begin{itemize}
    \item \textbf{Magnitude Hack} (Exp~033B): Wide 1$\times$12 with a learnable output scalar ($\alpha$, initialized at 3.0). If multi-hop is merely magnitude amplification, a scalar should suffice. Result: PPL~34.11; $\alpha$ converges to 5.04.
    \item \textbf{Asymmetric Depth} (Exp~034B): Per-layer hop allocation [2,1,2,1,1,4,4,2] based on causal KL knockout maps, using 29\% fewer total hops. Result: PPL~34.59---matching uniform-3-hop Deep with 29\% less expert compute.
    \item \textbf{Decoupled Routing} (Exp~035): Per-hop projection matrices that shatter the echo chamber ($\cos(\Delta h_0, \Delta h_1)$: $0.805 \to 0.049$, a 94\% reduction). Result: PPL~34.05.
\end{itemize}

Every intervention lands inside the same band.
The optimizer compensates for architectural constraints: critical layers absorb work from pruned layers (034B), orthogonal hop updates don't translate to better PPL than collinear ones (035), and a single scalar captures the multi-hop magnitude effect (033B).

\Paragraph{Phase dynamics.}
Training curves (Figure~\ref{fig:convergence}) reveal four phases: (1)~Dense leads at 0--10K (fast gradient flow, no routing overhead); (2)~Wide crosses Dense at $\sim$10K (centroids crystallize); (3)~Deep crosses Dense at $\sim$20K (credit assignment resolves); (4)~Convergence at 30K--50K (all curves flatten, gaps stabilize).
The convergence phase is the one that matters: transient advantages from early training vanish as all configurations find equally good minima.

\Paragraph{Implication for MoE design.}
Routing topology is quality-neutral; routing capacity is not; but routing mechanism barely matters.
At matched parameter budgets, the linear router's advantage shrinks to 1.2\%---comparable to the spread among cosine topology variants.
Investing extra parameters in routing (either cosine or linear) pays off; investing them in expert MLP width does not.
Cosine routing at $\dspace = 64$ saves 80\% of routing parameters at a cost of 0.38--1.17~PPL depending on whether the comparison is iso-parameter or not.

\subsection{Downstream Zero-Shot Evaluation}
\label{sec:downstream}

Zero-shot evaluation on HellaSwag, ARC-Easy, and WinoGrande (Exp~066) tests whether equifinality extends beyond perplexity.
Three cosine variants (Wide, Deep, Magnitude Hack) show small absolute accuracy differences (1.6--2.9 percentage points across benchmarks), but formal TOST equivalence at $\delta = 0.02$ (2~pp) passes for only 1 of 9 variant--benchmark pairs (Deep vs.\ Magnitude Hack on HellaSwag).
McNemar's paired tests reveal that Wide~1$\times$12---the variant with the \emph{best} PPL---is the \emph{worst} downstream performer on HellaSwag and ARC-Easy ($p < 0.01$ vs.\ both Deep and Magnitude Hack), reinforcing that PPL ranking does not predict downstream ranking.
The linear router (PPL~32.76) shows no downstream advantage over cosine variants despite its 1.2~PPL perplexity lead.
All models perform near chance at 76--84M parameters (HellaSwag~$\approx$25\%, WinoGrande~$\approx$50\%), limiting the discriminative power of these benchmarks at this scale; these results should be interpreted cautiously.
Per-benchmark accuracy breakdowns and McNemar contingency tables are provided in the supplementary experiment logs.
Because downstream benchmarks at this scale are only weakly discriminative, we turn to the main question: what mechanism makes routing variants converge to nearly the same quality?

\section{Why Routing Topology Converges}
\label{sec:mechanism}

The equifinality result demands a mechanistic explanation: \emph{why} do structurally different routing topologies converge to the same quality?
We present four probes that answer this question.

\Paragraph{Echo Chamber: Multi-Hop Updates Are Collinear.}
We measure $\cos(\Delta h_0, \Delta h_1)$ across 20{,}480 tokens to test whether multi-hop updates are compositional (orthogonal) or redundant (collinear).
With shared \texttt{proj\_in}, the overall cosine is \textbf{0.805}---hops perform \textbf{convergent magnitude amplification}, not orthogonal composition.
This is the mechanistic explanation for equifinality: multi-hop routing does not implement compositional reasoning.
It implements repeated amplification in approximately the same direction, which a single hop with a larger output scale can replicate (Exp~033B: PPL~34.11 with learned scalar $\alpha = 5.04$).

Decoupled per-hop projections (Exp~035) reduce the cosine to \textbf{0.049} (94\% reduction) and PPL improves from 34.62 to 34.05---but this still lands inside the same 0.79~PPL convergence band.
Even when forced to produce genuinely orthogonal updates, the architecture cannot escape the quality ceiling set by expert capacity.
This confirms: \emph{the ceiling is set by the expert pool, not by how tokens traverse it}.

\Paragraph{Frozen Routing: Re-Routing Provides Marginal Correction.}
To measure the contribution of hop-to-hop re-routing, we freeze expert selection: all hops use the initial position $\text{pos}_0$.
At the 20K-step checkpoint (PPL~268, underfitting regime), freezing degrades PPL by $+6.8\%$ with Jaccard collapsing from 0.102 to 0.996.
At convergence (50K steps, PPL~34.62), the same intervention degrades PPL by $+7.6\%$ (34.62 $\to$ 37.26)---but normal-mode Jaccard is already \textbf{0.970}, meaning the model has learned to select nearly identical experts at each hop even with dynamic re-routing.

\begin{table}[h]
\centering
\caption{Frozen routing at two training stages. At convergence, the model already selects 97\% identical experts across hops, yet freezing still degrades PPL by 7.6\%---re-routing provides a small but genuine correction.}
\label{tab:frozen}
\begin{tabular}{lcccc}
\toprule
Checkpoint & Normal PPL & Frozen PPL & $\Delta$ & Normal Jaccard \\
\midrule
20K steps (underfitting) & 268.60 & 286.76 & $+6.8\%$ & 0.102 \\
50K steps (converged) & 34.62 & 37.26 & $+7.6\%$ & 0.970 \\
\bottomrule
\end{tabular}
\end{table}

This reveals a nuanced picture: re-routing is \emph{causally relevant} (freezing hurts), but at convergence the model has learned to minimize hop-to-hop expert divergence.
The 3\% of expert changes that re-routing provides are disproportionately valuable.

\begin{figure}[t]
    \centering
    \includegraphics[width=\textwidth]{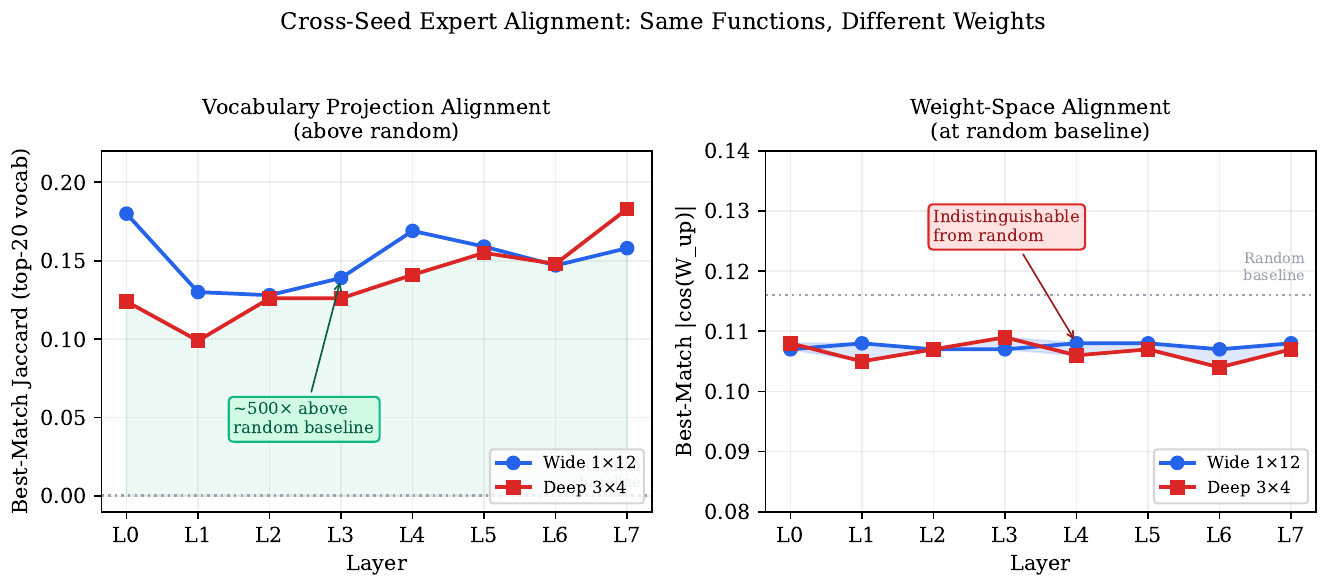}
    \caption{\textbf{Cross-Seed Expert Alignment.} Left: vocabulary projection Jaccard similarity (best-match) is ${\sim}500\times$ above random---experts develop partially overlapping functional roles across seeds. Right: raw $W_\text{up}$ cosine similarity is indistinguishable from random---these overlapping functions are realized through entirely different weight parameterizations. This divergence is the quantitative signature of equifinality.}
    \label{fig:cross_seed_align}
\end{figure}

\Paragraph{Cross-Seed Expert Stability: Different Weights, Same Functions.}
Comparing converged checkpoints from two seeds (42 and 137) for both Wide and Deep architectures, we measure expert alignment via (1) Jaccard similarity of top-20 projected vocabulary words (best-match across 1024 experts) and (2) cosine similarity of raw $W_\text{up}$ weight vectors.
Expert weights show \textbf{no cross-seed alignment}: best-match cosine similarity averages 0.107 for both architectures, indistinguishable from the random baseline (${\sim}0.116$ by extreme value theory for 1024 vectors in $\mathbb{R}^{1024}$).
However, vocabulary projections show \textbf{moderate functional alignment}: best-match Jaccard averages 0.15 (Wide) and 0.13 (Deep), far above the random baseline of ${\sim}0.0003$.
This pattern---random weight alignment, above-random functional alignment---confirms equifinality extends to the expert specialization level: different seeds and different topologies converge to functionally overlapping expert roles through entirely different parameterizations (Figure~\ref{fig:cross_seed_align}).

\Paragraph{Identity Falsification: Experts Are Computational, Not Copy-Paste.}
For all 8{,}192 experts, we compute $\cos(W_\text{down}, W_\text{up})$ to test whether experts are trivial copy-paste operations.
The overall mean cosine is \textbf{0.157}; \textbf{0.0\%} of experts are identity-like ($|\cos| > 0.8$) while \textbf{62.3\%} are near-orthogonal ($|\cos| < 0.2$).
Experts genuinely read one concept and write a different, orthogonal concept through SiLU nonlinearity---they are computational units, not lookup tables.
This rules out the hypothesis that equifinality is trivially caused by experts being too weak to support distinct trajectories.

\Paragraph{Summary.}
Multi-hop updates are collinear (echo chamber), routing converges to near-identical expert selection (frozen routing), different initializations produce functionally overlapping but weight-distinct expert pools (cross-seed stability), and experts are genuine nonlinear transformations (identity falsification).
Together, these explain why routing topology cannot determine quality: the expert pool sets a ceiling that no trajectory can exceed.
A companion paper \cite{controllability2026} shows that while routing \emph{topology} is interchangeable, individual expert \emph{identity} is causally meaningful---steering, suppression, and surgery on routing-identified experts produce large, selective output shifts.

\section{Redundancy Detection via Geometric Halting}
\label{sec:halting}

\subsection{Relative-Norm Halting}

The equifinality result (Section~\ref{sec:convergence}) establishes that routing topology does not determine quality.
A practical corollary: if multi-hop updates are largely redundant (as the echo chamber analysis confirms), we can \emph{detect and skip} redundant updates at inference without retraining.
This is not ``adaptive reasoning'' or ``anytime computation''---it is straightforward redundancy detection, enabled by the geometric routing substrate.

After each hop, if the relative update norm falls below a threshold $\varepsilon$:
\begin{equation}
    \frac{\|\Delta h\|}{\|x + h_\text{accum}\| + 10^{-6}} < \varepsilon \implies \text{halt}
\end{equation}
the token stops its trajectory early.
No retraining is required---this exploits the natural convergence behavior of expert updates.

\subsection{MLP Experts Enable Smooth Halting}

\begin{figure}[t]
    \centering
    \includegraphics[width=\textwidth]{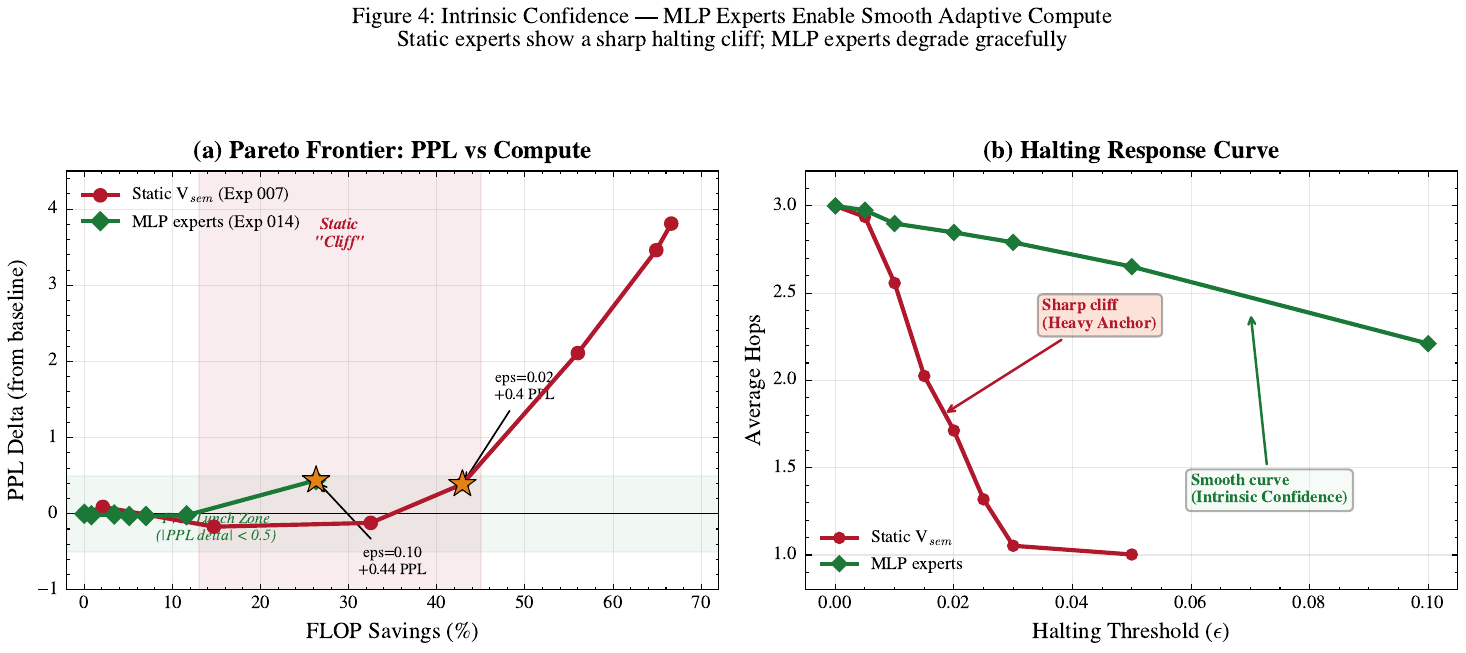}
    \caption{\textbf{Pareto Frontier.} (a)~MLP experts enable a smooth quality-compute tradeoff; static experts show a sharp cliff. (b)~Halting response: MLP average hops decrease gradually with $\varepsilon$, while static experts exhibit a phase transition.}
    \label{fig:pareto}
\end{figure}

With static experts (Exp~007), halting exhibits a \textbf{sharp phase transition} at $\varepsilon \approx 0.018$: average hops drop abruptly from 3.0 to 1.7.
This is the ``Heavy Anchor'' effect---$\|x\| \gg \|\Delta h\|$ for static vectors, so the relative norm is nearly identical across tokens, creating a cliff rather than a gradient.

With MLP experts (Exp~014), halting is \textbf{smooth and progressive}:

\begin{table}[h]
\centering
\caption{MLP halting Pareto points (Exp~014). Zero-shot on pre-trained model.}
\label{tab:halting}
\begin{tabular}{cccc}
\toprule
$\varepsilon$ & Avg Hops & FLOP Savings & $\Delta$PPL \\
\midrule
0.02 & 2.82 & 6.0\% & $-0.08\%$ \\
0.05 & 2.65 & 11.6\% & $+0.00\%$ \\
0.10 & 2.21 & 26.3\% & $+0.14\%$ \\
\bottomrule
\end{tabular}
\end{table}

At $\varepsilon = 0.05$, MLP halting saves 11.6\% of MoE FLOPs at \textbf{zero PPL cost}.
At $\varepsilon = 0.10$, it saves 26.3\% at only $+0.14\%$ PPL.

\Paragraph{Convergence validation (Exp~044c).}
We validate on the fully converged Deep~3$\times$4 model (50K~steps, PPL~34.62):

\begin{table}[h]
\centering
\caption{Halting at convergence (Exp~044c). The converged model is equally amenable to halting.}
\label{tab:halting_converged}
\begin{tabular}{cccc}
\toprule
$\varepsilon$ & Avg Hops & FLOP Savings & $\Delta$PPL \\
\midrule
0.02 & 2.58 & 14.2\% & $+0.00\%$ \\
0.05 & 2.44 & 18.8\% & $+0.03\%$ \\
0.10 & 2.25 & 25.0\% & $+0.12\%$ \\
0.15 & 2.08 & 30.7\% & $+0.43\%$ \\
\bottomrule
\end{tabular}
\end{table}

The converged model is \emph{more} amenable to halting: $\varepsilon = 0.02$ achieves 14.2\% FLOP savings at literally zero measurable PPL cost, and the headline operating point ($\varepsilon = 0.10$) saves 25\% at $+0.12\%$ PPL---consistent with the intermediate-checkpoint result.
This confirms halting is a stable architectural property, not an artifact of underfitting.

\Paragraph{Why zero-shot?}
The halting threshold is applied only at inference time.
The model was trained normally with all 3~hops.
This is critical: the model must \emph{learn to use all hops} before we can selectively skip those that contribute least.
As we show in Section~\ref{sec:greedy_horizon}, training with variable hop count produces the opposite effect---a model that never learns to use hops at all.

\Paragraph{Hardware validation.}
Profiling on RTX 3090 (Exp~017) confirms theoretical savings translate to real latency: halting reduces per-step latency from 4.32ms to 3.26ms ($-24.6\%$).
MoE routing overhead accounts for 21.5\% of CUDA time, suggesting custom Triton kernels could further improve throughput.

\section{Ablations}
\label{sec:ablations}

\subsection{Architecture Dualism: Macro vs Micro Regime}
\label{sec:dualism}

Our experiments reveal two fundamentally distinct operating modes for sparse MoE, which we term \emph{Architecture Dualism}:

\Paragraph{The Micro-Regime} ($M \geq 512$, expert\_size $= 1$): many tiny experts, best with Deep (multi-hop) routing, reaching PPL~304.50 at 36M.

\Paragraph{The Macro-Regime} ($M = 64$, expert\_size $\geq 8$): few large experts, best with Wide (single-hop) routing, reaching PPL~298.91 at 36M.

\begin{table}[h]
\centering
\caption{Architecture Dualism: optimal topology depends on expert granularity.}
\label{tab:dualism}
\begin{tabular}{lccccc}
\toprule
Regime & $M$ & Expert Size & Best Topology & PPL & vs Dense \\
\midrule
Micro & 512 & 1 & Deep ($3 \times 4$) & 304.50 & $-0.5\%$ \\
Macro & 64 & 8 & Wide ($1 \times 6$) & \textbf{298.91} & $\mathbf{-1.1\%}$ \\
\bottomrule
\end{tabular}
\end{table}

The Convergence Marathon (Section~\ref{sec:convergence}) validates this at scale: Wide 1$\times$12 (PPL~33.93) dominates raw quality while Deep 3$\times$4 (PPL~34.62) buys elastic compute (Section~\ref{sec:halting}).

\subsection{Centroid Density Determines Multi-Hop Quality}
\label{sec:nav_resolution}

The Architecture Dualism is explained by centroid density in routing space.
The decisive evidence comes from comparing two architectures at iso-FLOP:

\begin{table}[h]
\centering
\caption{Centroid density vs.\ per-hop rank: rank-4 with 512 centroids beats rank-16 with 128 centroids.}
\label{tab:paradox}
\begin{tabular}{lccccc}
\toprule
Architecture & $M$ & Expert Size & $\topk$/hop & Per-hop Rank & PPL \\
\midrule
Singleton Deep (Exp~012) & 512 & 1 & 4 & 4 & \textbf{304.50} \\
Granular Deep (Exp~022) & 128 & 4 & 4 & 16 & 337.26 \\
\bottomrule
\end{tabular}
\end{table}

A model with per-hop rank \textbf{4} outperforms one with per-hop rank \textbf{16} by 33~PPL points.
If per-hop bandwidth were the bottleneck, the rank-16 model should dominate.
Instead, the \emph{only} advantage of the rank-4 model is its \textbf{512 expert centroids} vs 128---a denser routing manifold that enables precise multi-hop navigation.

Table~\ref{tab:full_ablation} shows the complete pattern across granularity configurations:

\begin{table}[h]
\centering
\caption{Complete hop-width ablation at iso-FLOP (48 active neurons).}
\label{tab:full_ablation}
\begin{tabular}{lcccccr}
\toprule
Architecture & $M$ & es & Hops & $k$/hop & Hop Rank & PPL \\
\midrule
\multicolumn{7}{c}{\emph{Multi-hop (trajectory navigation)}} \\
Singleton Deep (Exp~012) & 512 & 1 & 3 & 4 & 4 & \textbf{304.50} \\
Macro Deep (Exp~018) & 64 & 8 & 3 & 2 & 16 & 337.02 \\
Hybrid (Exp~021) & 64 & 8 & 2 & 3 & 24 & 319.85 \\
Granular Deep (Exp~022) & 128 & 4 & 3 & 4 & 16 & 337.26 \\
\midrule
\multicolumn{7}{c}{\emph{Single-hop (parallel selection)}} \\
Macro Wide (Exp~020) & 64 & 8 & 1 & 6 & 48 & \textbf{298.91} \\
Granular Wide (Exp~023) & 128 & 4 & 1 & 12 & 48 & 304.79 \\
Dense d\_ff=48 (Exp~019) & --- & --- & --- & --- & 48 & 302.24 \\
\bottomrule
\end{tabular}
\end{table}

\subsection{The Greedy Horizon: Why Curriculum Halting Fails}
\label{sec:greedy_horizon}

Inference-time halting (Section~\ref{sec:halting}) achieves FLOP savings by exploiting a pre-trained model's natural convergence.
A natural question: can we train the model to be halting-aware from the start?
Inspired by stochastic depth \cite{huang2016deep}, we test a \textbf{trajectory curriculum}---progressively unlocking hop count during training:

\begin{center}
\begin{tabular}{lcc}
\toprule
Phase & Steps & max\_hops \\
\midrule
1 (Bootstrap) & 0--1000 & 1 always \\
2 (Unlock) & 1000--3000 & random $\{1, 2\}$ \\
3 (Full) & 3000--5000 & random $\{1, 2, 3\}$ \\
\bottomrule
\end{tabular}
\end{center}

\Paragraph{The result}: the model becomes \emph{lazy}---halting degradation is $26\times$ flatter, but only because hops 1--2 have become vestigial:

\begin{table}[h]
\centering
\caption{Trajectory Curriculum: the model becomes lazy (Exp~024).}
\label{tab:curriculum}
\begin{tabular}{lcccc}
\toprule
Condition & PPL@3-hop & PPL@2-hop ($\Delta$) & PPL@1-hop ($\Delta$) & FLOPs at 1-hop \\
\midrule
Baseline & \textbf{302.54} & 312.92 (+3.4\%) & 366.48 (+21.1\%) & $-67\%$ \\
Curriculum & 316.14 & 316.18 (+0.0\%) & 318.62 (+0.8\%) & $-67\%$ \\
\bottomrule
\end{tabular}
\end{table}

We term this \textbf{The Greedy Horizon}: when the network cannot \emph{guarantee} it will receive future hops, it rationally refuses to plan across them.
Distributing information across hops is a cooperative strategy that requires a guaranteed planning horizon.
The curriculum's stochastic truncation destroys this guarantee, causing the model to collapse into a myopic single-hop predictor.

\Paragraph{The lesson}: halting must be \textbf{zero-shot at inference only}.
Train with full depth so the model develops rich multi-hop trajectories.
Then apply inference-time halting (Section~\ref{sec:halting}) to skip those trajectories that happen to converge early.

\subsection{Sparsity Ratio and Temperature Interaction}
\label{sec:sparsity}

The baseline top-$K{=}12$ from $N{=}1024$ experts (1.17\% active) was chosen to match Deep 3$\times$4 (3 hops $\times$ 4 per hop), not from sparsity optimization.
We sweep $K \in \{4, 6, 8, 12, 24, 48\}$ at fixed $\tau{=}30$ (Wave~1), then test the $K \times \tau$ interaction at lower temperatures (Wave~2).

\begin{table}[h]
\centering
\caption{$K \times \tau$ ablation (Exp~051). Best PPL in \textbf{bold}. All runs: $N{=}1024$, Wide 1-hop, seed~42, 50K steps.}
\label{tab:ktau}
\begin{tabular}{lcccc}
\toprule
Config & $K$ & $\tau$ & Best PPL & $\Delta$ vs baseline \\
\midrule
Baseline & 12 & 30 & 33.93 & --- \\
$K{=}12$, $\tau{=}15$ & 12 & 15 & 34.06 & $+0.13$ \\
$K{=}24$, $\tau{=}30$ & 24 & 30 & 34.12 & $+0.19$ \\
$K{=}24$, $\tau{=}15$ & 24 & 15 & \textbf{33.85} & $\mathbf{-0.08}$ \\
$K{=}24$, $\tau{=}\text{learned}$ & 24 & 11--15 & 34.16 & $+0.23$ \\
$K{=}48$, $\tau{=}30$ & 48 & 30 & 33.92 & $-0.01$ \\
$K{=}48$, $\tau{=}10$ & 48 & 10 & 33.98 & $+0.05$ \\
\bottomrule
\end{tabular}
\end{table}

$K{=}24$ at $\tau{=}15$ achieves 33.85---the best single cosine-routing configuration---by redistributing routing mass so all 24 experts contribute meaningfully.
However, the 0.08~PPL improvement is within seed variance ($\sigma = 0.25$~PPL for Wide).
All 10 configurations fall within 0.82~PPL, extending the equifinality finding to the sparsity--temperature hyperparameter space.

\subsection{Composition Invariance}
\label{sec:composition}

The standard composition of rank-1 expert outputs is a weighted linear sum: $\Delta h = \sum_k w_k \cdot \text{SiLU}(h \cdot \mathbf{d}_k) \cdot \mathbf{u}_k$.
We test whether non-linear composition can increase expressiveness while preserving per-expert monosemanticity (Exp~056).

All na\"ive composition methods degrade quality: Cross-Expert Gating reaches PPL~$>$44 (killed at 17.5K steps), Grouped Gated reaches PPL~$>$62 (killed at 9.4K steps), and Grouped SiLU achieves PPL~34.92 ($+0.99$ vs baseline).
The failure pattern is clear: applying non-linearity across $\dmodel$ dimensions mixes expert directions, destroying the additive structure that rank-1 experts rely on.

This extends equifinality from routing topology to composition function: neither \emph{how} tokens select experts nor \emph{how} expert outputs are combined determines quality---only the experts themselves.
Follow-up work exploring activation-level gating (SE) and active compute scaling, which do improve quality by operating at the right granularity, is presented in a companion paper \cite{mol2026}.

\section{Discussion}
\label{sec:discussion}

\subsection{Topology Breadth}

Our equifinality claim covers five cosine-routing variants that share rank-1 expert pools, $K{=}12$ top-$k$ selection, and the same $d_\text{space}{=}64$ bottleneck.
Our $K \times \tau$ ablation (Section~\ref{sec:sparsity}) shows that varying sparsity within the same pool produces only 0.82~PPL variation (comparable to the topology band), but this is not the same as varying pool structure.

Exp~070 extends the equifinality test to rank-16 experts (256 experts, 4 layers, $K{=}4$).
Wide~1$\times$4 and Deep~2$\times$2 converge within 0.55~PPL ($37.72 \pm 0.08$ vs.\ $38.27 \pm 0.05$, 2 seeds each), comparable to the rank-1 gap of 0.69~PPL.
The higher absolute PPL is attributable to halved attention depth (4 vs.\ 8 layers) needed to match the ${\sim}84$M parameter budget, not a failure of the routing mechanism.
This addresses the concern that equifinality is an artifact of the rank-1 constraint, where experts are too limited for topology differences to express themselves.
Note that the topology pairs differ across rank settings (1$\times$4/2$\times$2 vs.\ 1$\times$12/3$\times$4), so cross-rank gap comparisons are suggestive rather than strictly controlled.

It is plausible that equifinality breaks down for architecturally more diverse comparisons---\eg, top-2 with 8 full-rank experts (Mixtral-style) vs.\ top-12 with 1024 rank-1 experts.
We view our result as establishing a lower bound on topology invariance: within the class of cosine-routed MoE at ranks 1 and 16, topology does not matter.

\Paragraph{Modality and task scope.}
Our equifinality claim is established for \emph{language modeling} perplexity.
Concurrent work on vision MoE \cite{promoe2026} reports that routing topology matters critically for diffusion transformers, where spatial redundancy and functional heterogeneity of visual tokens hinder expert specialization without explicit routing guidance.
Similarly, Chain-of-Experts \cite{xiao2025chainofexperts} shows compositional multi-hop benefits for structured math reasoning---a setting where our echo chamber analysis (collinear updates) may not apply.
Our Decoupled Routing experiment (Exp~035) tested the same principle (independent per-hop routing) for language modeling and found equifinality preserved, suggesting the distinction is task-driven rather than architectural.
Language tokens are semantically dense with pronounced inter-token variation, which may make routing topology less critical than for spatially redundant visual patches or structured mathematical expressions.

\Paragraph{Post-training routing alignment.}
Zhou et al. \cite{roma2025} show a 10--20\% accuracy gap between existing MoE routers and oracle routing in post-training, with routing manifold alignment improving generalization by 5.5--8.6\%.
This does not contradict pre-training equifinality: the gap measures how far current routers are from an oracle upper bound, not whether different routing topologies converge to the same quality.
A suboptimal router can still be topology-invariant---multiple topologies may converge to the same suboptimal point relative to the oracle.
Nevertheless, this suggests that routing \emph{efficiency} (how quickly the ceiling is reached) may depend on routing design even when the ceiling itself is topology-invariant.

\subsection{Multi-Epoch Training and Memorization}

Our marathon configuration trains for ${\sim}14$ epochs on WikiText-103 (1.64B tokens / 117M corpus tokens).
A natural concern is memorization.
We address this with three observations.
First, validation loss is \emph{still decreasing} at step 50K for all models: Wide drops $-0.013$ loss over the last 5 checkpoints (steps 40K--50K), Deep drops $-0.012$, and Dense drops $-0.013$.
Second, the train--validation gap at convergence is small: $+0.13$ loss for Wide (3.39 train vs.\ 3.52 val) and $+0.15$ for Deep---comparable to the $+0.02$ gap for the Dense baseline, which has $93\times$ more active parameters.
Third, our claim is \emph{comparative}: equifinality measures differences \emph{between} models trained under identical data conditions; any memorization affects all variants equally and cancels in paired comparisons.
Cross-corpus replication on OpenWebText (Exp~067) confirms generalization: with 3 seeds per condition (6 runs total), the Wide--Deep gap is just 0.03~PPL between means (Wide $34.25 \pm 0.20$, Deep $34.22 \pm 0.02$)---$6\times$ smaller than seed noise.
Training on OpenWebText uses ${\sim}6.5$ epochs over 507M tokens from a different domain (web text vs.\ Wikipedia), and the equifinality finding cannot be attributed to memorizing WikiText-103 specifically.
The variance structure itself replicates across corpora ($\sigma_\text{wide} \approx 0.20$--$0.25$, $\sigma_\text{deep} \approx 0.02$--$0.03$ on both).
Replication on a truly single-epoch corpus would further strengthen the result.

\subsection{Scale Limitations}

All experiments use WikiText-103 at 76--138M parameters (138M includes the $K{=}128$ active-compute scaling experiment, Exp~055).
While the equifinality finding is statistically robust at this scale (TOST equivalence, multi-seed validation), we cannot claim it generalizes to billion-parameter models or diverse training corpora without direct evidence.
Fine-grained MoE scaling laws \cite{ludziejewski2024scaling} suggest that granularity benefits persist at scale, but the interaction between routing topology and scale remains untested for our architecture.
Our 1.2B-parameter configuration (Exp~046, in preparation) targets this gap.

\subsection{Statistical Coverage}

Our equivalence claim rests on paired bootstrap confidence intervals and TOST with an equivalence margin of $\delta = 0.03$ cross-entropy loss ($\approx$1~PPL).
This margin is justified by comparison to intra-seed variance (0.013 loss for Wide 1$\times$12 across seeds) and to the linear router's clearly-outside-band advantage (0.033 loss).
Block bootstrap with block sizes 5 and 10 (accounting for sequential batch correlation in WikiText-103) produces CIs that are at most $1.09\times$ wider than independent bootstrap (Appendix~\ref{app:block_bootstrap}), confirming that batch correlation does not inflate our equivalence claims.
Full multi-seed validation (3 seeds $\times$ 5 variants $= 15$ runs) confirms all pairwise equivalences hold.

\subsection{Expert Contribution Magnitude}

One might hypothesize that equifinality is trivially explained by experts being too weak to influence the output.
Direct measurement refutes this: expert update norms average 39.6\% of attention output norms for Wide 1$\times$12 (14.5\% in layer~7 to 72.0\% in layer~3), and 148.2\% for Deep 3$\times$4.
Expert-zeroing ablation collapses PPL from 33.9 to 494.5 for Wide and from 34.6 to 846.9 for Deep---a 14--24$\times$ degradation.
Experts are substantial, load-bearing contributors; equifinality reflects genuine underdetermination of routing topology, not expert irrelevance.

\subsection{Deployment Considerations}

At this scale, the sequential expert loop in cosine routing is 1.8$\times$ slower per step than an iso-parameter dense FFN (3.5 vs.\ 1.9 steps/s).
Standard MoE with linear routers and parallel expert selection remains hardware-optimal for throughput on modern accelerators---and achieves better quality when routing parameter budget is unconstrained (Exp~036).
The practical value of cosine routing is not raw performance but the transparency properties it enables: geometric halting, centroid-based expert discovery, and routing-space interventions \cite{controllability2026}.

\section{Related Work}
\label{sec:related}

\Paragraph{Sparse Mixture of Experts.}
Sparse MoE was introduced by Shazeer et al. \cite{shazeer2017outrageously} with top-2 gating.
GShard \cite{lepikhin2021gshard} scaled to 600B parameters.
Switch Transformer \cite{fedus2022switch} simplified to top-1 with load balancing.
ST-MoE \cite{zoph2022stmoe} established stable training recipes.
Mixtral \cite{jiang2024mixtral} demonstrated competitive quality with 8 experts at top-2.
OLMoE \cite{muennighoff2025olmoe} scales fine-grained MoE to 6.9B parameters with 64 experts per layer (top-8), finding that router decisions saturate within 1\% of pretraining and that finer granularity consistently improves quality; their observational routing analysis documents expert specialization at scale but does not test whether alternative routing topologies yield equivalent quality.
Ludziejewski et al. \cite{ludziejewski2024scaling} derived scaling laws for fine-grained MoE, showing that smaller, more numerous experts consistently improve the loss--compute Pareto frontier up to a granularity of $G{=}16$ at the 1B scale; our rank-1 design pushes granularity far beyond their tested range.
Concurrent MoE scaling laws derive loss predictors over structural factors---data, parameters, active experts, granularity, and shared expert ratio---from hundreds of controlled experiments \cite{zhao2025comprehensive, tian2025leverage}; crucially, none include routing mechanism or topology as a factor, treating routing as fixed infrastructure.
Chaudhari et al. \cite{chaudhari2026moelens} show that in DeepSeekMoE, a single top expert's output has cosine similarity 0.95 with the full 6-expert ensemble, with only 5\% perplexity increase when using one expert---strong evidence that expert contributions are largely redundant.
Wang et al. \cite{wang2025buddymoe} exploit this redundancy for inference: substituting cached ``buddy'' experts with similar functionality yields negligible accuracy loss at 10\% throughput improvement, confirming that expert identity is interchangeable at the functional level.
All use learned router networks; we eliminate the router entirely.

\Paragraph{Router-free routing.}
Hash layers \cite{roller2021hash} use fixed hash functions, while DEMix \cite{gururangan2022demix} routes by domain metadata.
PEER \cite{he2024mixture} scales to $>$1M micro-experts via Product Key Memory, but retains a parameterized retrieval index.
ReMoE \cite{wang2025remoe} replaces top-$K$ with fully differentiable ReLU routing, improving scalability with expert count; RMoE \cite{qiu2025rmoe} adds GRU-based cross-layer recurrence to routing decisions.
Both alter routing \emph{mechanisms} but evaluate a single topology; our work is complementary, showing that routing topology is quality-neutral regardless of whether the routing function is cosine, linear, or ReLU.
Nguyen et al. \cite{nguyen2025statistical} prove that cosine and linear routers achieve the same regression convergence rate $O(\sqrt{\log(n)/n})$, providing theoretical support for our empirical finding that the mechanism gap is modest (${\sim}1.2\%$); their analysis also shows that norm perturbation restores polynomial parameter identifiability---our temperature $\tau{=}30$ may serve an analogous role.
DirMoE \cite{vahidi2026dirmoe} disentangles expert selection (Bernoulli) from contribution weighting (Dirichlet), achieving fully differentiable routing that matches or exceeds standard top-$K$---yet another routing mechanism producing equivalent quality.
Grouter \cite{xu2026grouter} takes the extreme approach of distilling routing structures from a trained MoE and using them as \emph{frozen} routers for a target model, finding that decoupled, preemptive routing \emph{improves} convergence over jointly-learned routing---evidence that expert weights absorb any fixed routing signal.
Our approach uses the token's own semantic representation via cosine similarity against learned centroids, requiring 80\% fewer routing parameters than a standard linear router and enabling geometric halting criteria unavailable to parameterized routers.

\Paragraph{Multi-hop and iterative computation.}
Chain-of-Experts \cite{xiao2025chainofexperts} applies sequential routing through expert chains, reporting compositional benefits for math reasoning (6.7\% loss reduction); however, this uses independent per-iteration routers and targets structured reasoning rather than general language modeling---our Decoupled Routing experiment (Exp~035), which similarly forces orthogonal per-hop updates, finds equifinality preserved for language modeling PPL.
Mixture of Depths \cite{raposo2024mixture} allows tokens to skip layers.
Universal Transformers \cite{dehghani2019universal} apply shared layers iteratively with adaptive depth.
\stmoe{} combines multi-hop routing with semantic re-navigation, where each hop's routing is informed by accumulated expert updates.

\Paragraph{Routing convergence and expert specialization.}
Aquino-Michaels \cite{aquinomichaels2026absorption} provides the most direct mechanistic support for equifinality: in sparse attention, learned soft gating differs from \emph{frozen random gates} by only 1.10~PPL (2.2\%) at 31M parameters, and at 1.7B scale the gap vanishes entirely (8.80 vs.\ 8.80~PPL).
The explanation is \emph{routing absorption}: the 80:1 parameter asymmetry between model weights and gate parameters means experts continuously co-adapt to compensate for whatever routing mask is imposed.
Tran et al. \cite{tran2025lmc} show that independently trained MoE models with the same architecture land in the same loss basin under permutation alignment (linear mode connectivity), establishing \emph{within-topology} convergence; our equifinality finding extends this to \emph{cross-topology} convergence, showing that even structurally different routing topologies converge to equivalent quality.
Wang et al. \cite{wang2026illusion} find that 2--5 domain-invariant ``standing committee'' experts capture 14--70\% of routing mass regardless of input domain, providing observational evidence for why topology may not matter: the same core experts dominate across conditions.
The SD-MoE authors \cite{sdmoe2026} decompose expert parameters into shared low-rank subspaces plus unique orthogonal complements, finding that experts share highly overlapping dominant spectral components---consistent with our echo chamber observation that multi-hop updates are collinear.
Li et al. \cite{li2026understanding} decompose MoE routing into cross-layer contributions, finding entanglement patterns that parallel our within-layer collinearity ($\cos(\Delta h_0, \Delta h_1) = 0.805$).
Guo et al. \cite{guo2025specialization} show that standard load-balancing losses produce expert overlap and overly uniform routing; orthogonality losses improve specialization by up to 23.8\%, suggesting the expert pool ceiling can be raised by diversifying expert functions rather than changing routing topology.

\Paragraph{Adaptive depth and stochastic depth.}
Stochastic depth \cite{huang2016deep} randomly drops layers during training for regularization.
Our trajectory curriculum (stochastic hop depth) applies the same idea to MoE hops---but yields a negative result (Section~\ref{sec:greedy_horizon}), demonstrating that multi-hop MoE requires guaranteed depth during training.

\section{Conclusion}
\label{sec:conclusion}

At the scales we test, routing topology changes little about final language-model quality.
Our central finding is negative: \textbf{within the tested regime, routing topology does not determine asymptotic language modeling quality}.
Five cosine-routing configurations are statistically equivalent within 1~PPL (TOST, all 10 pairs $p < 0.05$, 15 runs), the gap to a standard linear router is mostly explained by routing capacity, and multi-hop updates are collinear rather than compositional.
A companion paper \cite{controllability2026} demonstrates that while routing \emph{topology} is interchangeable, individual expert \emph{identity} is causally meaningful.

\Paragraph{Practical payoff.}
Since routing topology is quality-neutral, we exploit geometric routing for compute efficiency: zero-shot halting saves 25\% of MoE FLOPs at $+0.12\%$ PPL, and static pruning removes 29\% of expert computations from low-impact layers with no quality loss.

\Paragraph{Limitations.}
Primary results are on WikiText-103 at 76--138M parameters.
Supporting evidence extends to rank-16 experts (0.55~PPL gap), zero-shot downstream benchmarks, and OpenWebText cross-corpus replication (0.03~PPL gap, 6 runs), but downstream accuracy remains near-chance at this scale and rank-16 topology differs from rank-1.
We do not claim generalization to billion-parameter scale without direct evidence.
Concurrent work on routing absorption \cite{aquinomichaels2026absorption} suggests equifinality may strengthen with scale, but hash routing degradation widens at higher expert counts, so the interaction may differ across routing mechanisms.
Equifinality may not hold for vision MoE \cite{promoe2026} or structured reasoning tasks (Section~\ref{sec:discussion}).

\Paragraph{Design implication.}
This shifts the MoE design question from ``Which topology is best?'' to ``Which routing substrate is simplest, cheapest, and most useful for the behaviors we care about?''
The answer depends on the deployment context: if throughput is paramount, standard linear routing with parallel expert selection remains optimal; if inference-time adaptivity or routing transparency matter, geometric routing provides unique capabilities (halting, centroid-based discovery, routing-space interventions) at modest quality cost.
More broadly, our results suggest that research effort is better spent on expert capacity and routing efficiency than on increasingly elaborate routing topologies.

\bibliographystyle{unsrt}
\bibliography{references}

\appendix

\section{Training Configuration}
\label{app:training}

Table~\ref{tab:training_config} provides all hyperparameters used in the convergence marathon (Exps~025--027) and all subsequent experiments.

\begin{table}[h]
\centering
\caption{Complete training configuration for Marathon-scale experiments.}
\label{tab:training_config}
\begin{tabular}{ll}
\toprule
\textbf{Parameter} & \textbf{Value} \\
\midrule
\multicolumn{2}{l}{\emph{Architecture}} \\
$\dmodel$ & 1024 \\
Attention heads & 16 (head dim = 64) \\
Layers & 8 \\
Experts per layer ($M$) & 1024 \\
Expert size (rank) & 1 \\
$\dspace$ (routing dimension) & 64 \\
Temperature ($\tau$) & 30 (fixed) \\
Vocabulary & 32,000 BPE \\
Max sequence length & 1024 \\
Position encoding & Rotary (RoPE) \\
Weight tying & Embedding $\leftrightarrow$ LM head \\
\midrule
\multicolumn{2}{l}{\emph{Training}} \\
Optimizer & AdamW ($\beta_1{=}0.9$, $\beta_2{=}0.95$, $\epsilon{=}10^{-8}$) \\
Learning rate & $3 \times 10^{-4}$ \\
LR schedule & Cosine decay to $0.1\times$ \\
Warmup steps & 1,000 \\
Training steps & 50,000 \\
Batch size & 8 sequences $\times$ 512 tokens \\
Gradient accumulation & 8 steps \\
Effective batch & 32,768 tokens/step \\
Total tokens & 1.64B (${\sim}$14 epochs) \\
Weight decay & 0.01 \\
Gradient clipping & 1.0 (max norm) \\
Balance loss ($\alpha$) & 0.05 \\
Dropout & 0.1 \\
Precision & bfloat16 (mixed) \\
\midrule
\multicolumn{2}{l}{\emph{Hardware}} \\
GPU & NVIDIA RTX 3090 (24~GB) \\
Wall-clock per marathon run & 7--9h \\
\bottomrule
\end{tabular}
\end{table}

\section{Statistical Validation Details}
\label{app:statistics}

\subsection{Paired Bootstrap Confidence Intervals}

For each pair of cosine routing variants $(i, j)$, we compute per-batch loss differences $\delta_b = \ell_{i,b} - \ell_{j,b}$ for $b = 1, \ldots, 50$ validation batches.
We resample these differences 10{,}000 times with replacement (seed = 42) and report the 2.5th and 97.5th percentiles as the 95\% CI.
This paired design cancels batch-level variance, yielding CIs approximately $10\times$ tighter than unpaired bootstrap.

\begin{table}[h]
\centering
\caption{Paired bootstrap 95\% CIs on per-batch loss differences (all cosine variant pairs).}
\label{tab:paired_ci}
\begin{tabular}{lccc}
\toprule
Pair & Mean $\Delta$ Loss & 95\% CI & Zero in CI? \\
\midrule
Wide vs Deep & $-0.023$ & [$-0.031$, $-0.015$] & No \\
Wide vs Magnitude & $-0.004$ & [$-0.010$, $+0.003$] & Yes \\
Wide vs Asymmetric & $-0.020$ & [$-0.027$, $-0.012$] & No \\
Wide vs Decoupled & $-0.007$ & [$-0.016$, $+0.001$] & Yes \\
Deep vs Magnitude & $+0.020$ & [$+0.012$, $+0.026$] & No \\
Deep vs Asymmetric & $+0.004$ & [$-0.004$, $+0.011$] & Yes \\
Deep vs Decoupled & $+0.016$ & [$+0.008$, $+0.024$] & No \\
Mag vs Asymmetric & $-0.016$ & [$-0.023$, $-0.009$] & No \\
Mag vs Decoupled & $-0.004$ & [$-0.011$, $+0.004$] & Yes \\
Asym vs Decoupled & $+0.012$ & [$+0.005$, $+0.020$] & No \\
\bottomrule
\end{tabular}
\end{table}

Six of ten pairs show statistically significant differences (zero outside CI), but all differences are small ($|\Delta| \leq 0.023$ loss $\approx 0.75$~PPL).

\subsection{TOST Equivalence Test}

The Two One-Sided Tests (TOST) procedure \cite{schuirmann1987comparison} tests $H_0$: $|\mu_i - \mu_j| \geq \delta$ against $H_1$: $|\mu_i - \mu_j| < \delta$.
We reject $H_0$ (declare equivalence) if the entire 90\% CI on the paired difference falls within $(-\delta, +\delta)$.

\begin{table}[h]
\centering
\caption{TOST equivalence results at three margins.}
\label{tab:tost}
\begin{tabular}{lcc}
\toprule
Margin ($\delta$) & Pairs Equivalent (of 10) & Interpretation \\
\midrule
0.02 loss ($\approx 0.6$~PPL) & 5/10 & Strict: half equivalent \\
\textbf{0.03 loss ($\approx 1.0$~PPL)} & \textbf{10/10} & \textbf{All equivalent} \\
0.05 loss ($\approx 1.6$~PPL) & 10/10 & All equivalent (generous) \\
\bottomrule
\end{tabular}
\end{table}

\subsection{Seed Variance}

\begin{table}[h]
\centering
\caption{Intra-seed variance (paired bootstrap, seeds 42 vs 137).}
\label{tab:seed_var}
\begin{tabular}{lccc}
\toprule
Configuration & Seed 42 vs 137 & 95\% CI & Significant? \\
\midrule
Wide 1$\times$12 & $-0.013$ & [$-0.021$, $-0.005$] & Yes \\
Deep 3$\times$4 & $-0.003$ & [$-0.010$, $+0.005$] & No \\
\bottomrule
\end{tabular}
\end{table}

\subsection{Linear Router as Negative Control}

The linear router (Exp~036) serves as a negative equivalence control.
Its mean loss is 0.033 below the best cosine variant (Wide): 90\% CI = [$-0.041$, $-0.025$], entirely below zero.
This confirms the equivalence band is specific to cosine routing variants and is not an artifact of insensitive measurement.

\subsection{Block Bootstrap Robustness}
\label{app:block_bootstrap}

WikiText-103 validation batches are drawn sequentially, introducing potential autocorrelation.
Block bootstrap \cite{schuirmann1987comparison} accounts for this by resampling contiguous blocks rather than individual batches.
We repeat the TOST analysis with block sizes $b = 5$ and $b = 10$ (10{,}000 resamples each).

\begin{table}[h]
\centering
\caption{Block bootstrap TOST equivalence ($\delta = 0.03$ loss). CIs are at most $1.09\times$ wider than independent bootstrap; all conclusions are preserved.}
\label{tab:block_tost}
\begin{tabular}{lccc}
\toprule
Pair & Standard & Block ($b{=}5$) & Block ($b{=}10$) \\
\midrule
Wide vs Deep    & Equiv. & Equiv. & Equiv. \\
Wide vs Mag     & Equiv. & Equiv. & Equiv. \\
Wide vs Asym    & Equiv. & Equiv. & Equiv. \\
Wide vs Decoup  & Equiv. & Equiv. & Equiv. \\
Deep vs Mag     & Equiv. & Equiv. & Equiv. \\
Deep vs Asym    & Equiv. & Equiv. & Equiv. \\
Deep vs Decoup  & Equiv. & Equiv. & Equiv. \\
Mag vs Asym     & Equiv. & Equiv. & Equiv. \\
Mag vs Decoup   & Equiv. & Equiv. & Equiv. \\
Asym vs Decoup  & Equiv. & Equiv. & Equiv. \\
\midrule
Total & 10/10 & 10/10 & 10/10 \\
\bottomrule
\end{tabular}
\end{table}

The widest pair (Wide vs Deep, $\Delta = -0.023$) has 90\% CIs of [$-0.030$, $-0.017$] (standard), [$-0.031$, $-0.019$] (block-5), and [$-0.030$, $-0.019$] (block-10)---all within $[-0.03, +0.03]$.
Note that the standard bootstrap lower bound ($-0.030$) touches the equivalence boundary exactly; TOST uses the closed interval $[-\delta, +\delta]$, so this pair passes, but it is the tightest margin in the analysis.
Maximum CI width expansion is $1.09\times$ at $b{=}5$, confirming that sequential correlation does not materially affect our equivalence conclusions.

\section{Underfitting-Regime Results}
\label{app:underfitting}

These results are from models trained for 5--20K steps, well before convergence.
We include them as mechanistic characterization of how expert depth interacts with training stage.
The convergence results (Section~\ref{sec:convergence}) are the authoritative quality measurements.

\subsection{The Linear Trap}

With static expert vectors $V_i$, the multi-hop update collapses to a linear sum regardless of trajectory.
In Exp~006, static Deep (3~hops $\times$ top-4) beats static Wide (1~hop $\times$ top-12) by only $+0.4\%$ PPL (321.43 vs 322.65).
Depth barely matters when experts cannot compose nonlinearly.

\subsection{MLP Experts Unlock Depth}

Replacing static vectors with rank-1 MLP experts (Eq.~\ref{eq:mlp_expert}) yields $-4.4\%$ PPL (320.64 $\to$ 306.60) at only $+4.4\%$ parameter cost (Exp~010).
Trajectory displacement increases $3.5\times$.

\subsection{Transient Depth Advantage}

\begin{figure}[t]
    \centering
    \includegraphics[width=\textwidth]{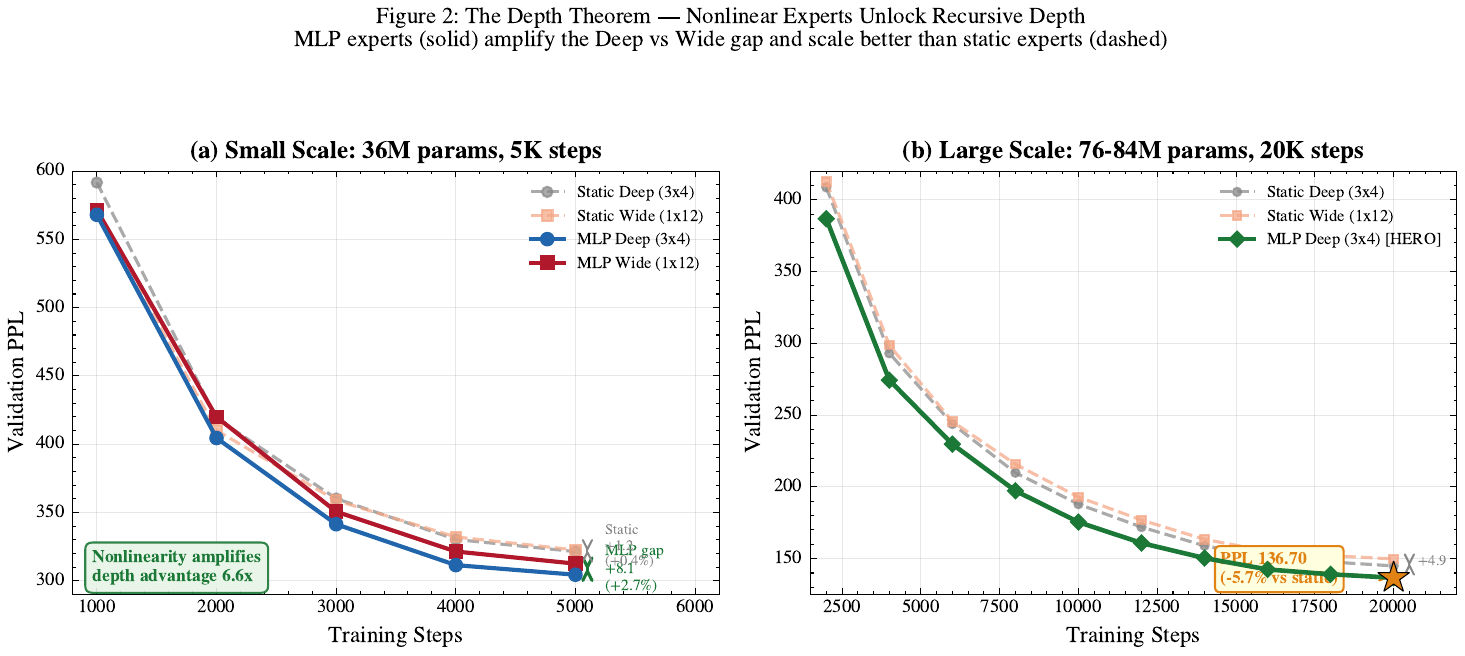}
    \caption{\textbf{Depth advantage during underfitting.} At 36M (a) and 76M (b), MLP experts amplify the Deep-vs-Wide gap. However, this advantage \emph{vanishes at convergence} (Section~\ref{sec:convergence}): at 50K steps, Wide (PPL~33.93) beats Deep (PPL~34.62).}
    \label{fig:depth_theorem}
\end{figure}

\begin{table}[t]
\centering
\caption{Depth advantage with static vs MLP experts at two scales.}
\label{tab:depth}
\begin{tabular}{lcccc}
\toprule
Expert Type & Scale & Deep PPL & Wide PPL & Depth Advantage \\
\midrule
Static $V_\text{sem}$ & 36M & 321.43 & 322.65 & $+0.4\%$ \\
MLP (rank-1) & 36M & 304.50 & 312.61 & $+2.7\%$ \\
\midrule
Static $V_\text{sem}$ & 76M & 144.95 & 149.80 & $+3.3\%$ \\
MLP (rank-1) & 76M & \textbf{136.70} & 149.80 & $+8.8\%$ \\
\bottomrule
\end{tabular}
\end{table}

At 36M scale, MLP Deep (PPL~304.50) beats MLP Wide (312.61) by $2.7\%$---a $6.6\times$ amplification over the static baseline's $0.4\%$.
At 76M scale, the advantage grows to $8.8\%$.

\begin{figure}[t]
    \centering
    \includegraphics[width=\textwidth]{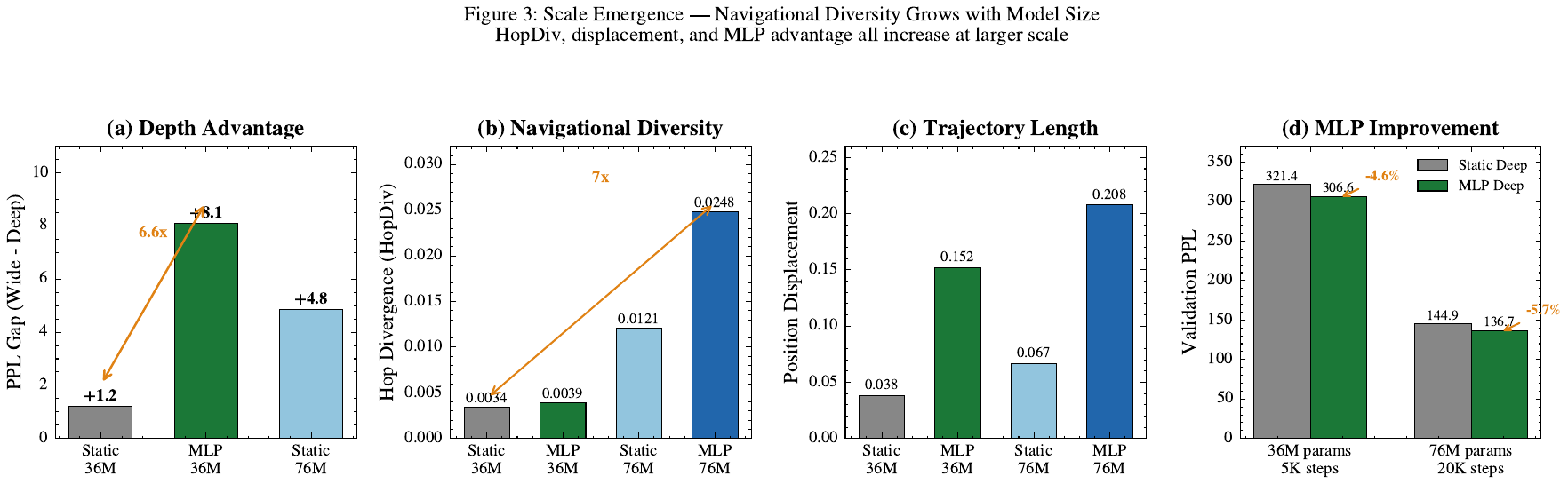}
    \caption{\textbf{Scale Emergence.} (a)~Depth advantage grows with scale. (b)~Navigational diversity (HopDiv) increases $7\times$ from 36M to 76M. (c)~Trajectory displacement grows. (d)~MLP improvement compounds: $-4.6\%$ at 36M $\to$ $-5.7\%$ at 76M.}
    \label{fig:scale_emergence}
\end{figure}

However, at convergence (50K steps, Section~\ref{sec:convergence}), the depth advantage \emph{inverts}---Wide beats Deep by $-2.0\%$.
The underfitting-regime depth advantage is transient, not a scaling law.

\section{Experiment Index}
\label{app:experiments}

\begin{table}[h]
\centering
\caption{Selected experiment index (62 total). Experiments cited to Ternovtsii and Bilak \cite{controllability2026} are detailed in the companion paper. Full logs in supplementary materials.}
\label{tab:exp_index}
\small
\begin{tabular}{llll}
\toprule
Exp & Description & Key Result & Section \\
\midrule
001--003 & Routing mode ablation & Semantic re-route best & Section~\ref{sec:method} \\
004--009 & Architecture search & $\dspace{=}64$ optimal & Section~\ref{sec:method} \\
010 & MLP expert introduction & $-4.4\%$ PPL & App.~\ref{app:underfitting} \\
012--013 & Scale emergence & Depth $+8.8\%$ at 76M & App.~\ref{app:underfitting} \\
014, 017 & Halting + profiling & 25\% FLOPs saved & Section~\ref{sec:halting} \\
018--023 & Hop-width ablation & Centroid density key & Section~\ref{sec:dualism} \\
024 & Trajectory curriculum & Greedy Horizon failure & Section~\ref{sec:greedy_horizon} \\
025--027 & Convergence marathon & Equifinality band & Section~\ref{sec:convergence} \\
029--032 & Causal interventions & KL up to 0.47 & \cite{controllability2026} \\
033--035 & Routing interventions & All within band & Section~\ref{sec:convergence} \\
036 & Linear router control & PPL 32.76 ($5.3\times$ params) & Section~\ref{sec:convergence} \\
037--039 & Expert controllability & Median $+321\%$ steering & \cite{controllability2026} \\
042--043 & Multi-seed (Wide/Deep, 2 seeds) & 0.74~PPL window & Section~\ref{sec:convergence} \\
044a--b & Cross-routing controllability & Linear also steerable & \cite{controllability2026} \\
044c & Halting at convergence & 25\% FLOPs saved at $+0.12\%$ PPL & Section~\ref{sec:halting} \\
045 & Routing statistics & Gini gradient L0--L7 & \cite{controllability2026} \\
047 & Norm diagnostic & Experts load-bearing & Section~\ref{sec:discussion} \\
048 & Bootstrap CI / TOST & All 10 pairs equivalent & Section~\ref{sec:convergence} \\
049 & Multi-seed (5 variants $\times$ 3 seeds) & 0.79~PPL band, 15 runs & Section~\ref{sec:convergence} \\
050 & Steering robustness + 8 categories & Median $+321\%$, 98\% positive & \cite{controllability2026} \\
051 & $K \times \tau$ sparsity ablation & 0.82~PPL across 10 configs & Section~\ref{sec:sparsity} \\
052--053 & Grammar specialization + Zipf control & Syntax wins 8/8 layers & \cite{controllability2026} \\
056 & Na\"ive composition ablation & All variants worse & Section~\ref{sec:composition} \\
065 & Broader topology (hash, random, top-1) & 1.1--2.2~PPL from learned & Section~\ref{sec:convergence} \\
066 & Downstream zero-shot eval & PPL rank $\neq$ downstream rank & Section~\ref{sec:downstream} \\
067 & OpenWebText replication (multi-seed) & 0.03~PPL gap (6 runs, 3 seeds each) & Section~\ref{sec:discussion} \\
070 & Rank-16 equifinality & 0.55~PPL Wide--Deep gap & Section~\ref{sec:discussion} \\
\bottomrule
\end{tabular}
\end{table}

\end{document}